\documentclass[11pt]{article}

\usepackage[preprint]{acl}

\usepackage{times}
\usepackage{latexsym}

\usepackage[T1]{fontenc}

\usepackage[utf8]{inputenc}

\usepackage{microtype}

\usepackage{inconsolata}

\usepackage{graphicx}

\usepackage{subcaption}
\usepackage{caption}
\usepackage[utf8]{inputenc}
\usepackage{array}
\usepackage{tabularx}
\usepackage{xcolor}
\usepackage{booktabs}
\usepackage[font=small,skip=5pt]{caption}
\usepackage{enumitem}
\usepackage{makecell}
\usepackage{multirow}
\usepackage{xcolor}  
\usepackage{titlesec}
\usepackage[most]{tcolorbox} 
\usepackage{xcolor}
\usepackage{lipsum} 
\usepackage{algorithmic}
\usepackage{algorithm}

\definecolor{titlebg}{RGB}{60,60,60}   
\definecolor{boxbg}{RGB}{245,245,245}  

\usepackage{amsmath}
\usepackage{enumitem}
\usepackage{graphicx}

\usepackage{amssymb}

\usepackage{adjustbox}
\newcommand{\compacteqsetup}{%
  \small
}

\AtBeginEnvironment{equation}{\compacteqsetup}
\AtBeginEnvironment{equation*}{\compacteqsetup}

\usepackage{amsmath,amssymb,amsthm}

\newtheorem{assumption}{Assumption}
\newtheorem{proposition}{Proposition}
\newtheorem{lemma}{Lemma}

\title{MulFeRL: Enhancing Reinforcement Learning with Verbal Feedback in a Multi-turn Loop}

\author{
  \textbf{Xuancheng Li\textsuperscript{1,2,$\dagger$}},
  \textbf{Haitao Li\textsuperscript{1,2,$\dagger$}},
  \textbf{Yujia Zhou\textsuperscript{1,2}},
  \textbf{Yiqun Liu\textsuperscript{1}},
  \textbf{Min Zhang\textsuperscript{1}},
  \textbf{Qingyao Ai\textsuperscript{2,1,*}} \\
  \textsuperscript{1}Department of Computer Science and Technology, Tsinghua University, Beijing, China \\
  \textsuperscript{2}Quancheng Laboratory \\
  \texttt{lixuancheng23@mails.tsinghua.edu.cn} \\
  \texttt{liht22@mails.tsinghua.edu.cn, zhouyujia@mail.tsinghua.edu.cn} \\
  \texttt{yiqunliu@tsinghua.edu.cn, aiqy@tsinghua.edu.cn}
}

\begin{document}
\maketitle 
\begingroup
\renewcommand\thefootnote{\fnsymbol{footnote}}
\footnotetext[2]{Equal contribution.}
\footnotetext[1]{Corresponding author.}
\endgroup
\begin{abstract}
Reinforcement Learning with Verifiable Rewards (RLVR) is widely used to improve reasoning across domains, but outcome-only scalar rewards are often sparse and uninformative, especially for failed samples, where they indicate only failure without explaining why the reasoning breaks down. In this paper, we leverage richer verbal feedback to guide RLVR on failed samples and convert feedback-induced progress into trainable learning signals. 
We propose \textbf{MulFeRL} (\textbf{Mul}ti-turn \textbf{Fe}edback-guided \textbf{R}einforcement \textbf{L}earning), a multi-turn, event-triggered RLVR framework that combines \emph{progress induction} for feedback-guided regeneration of failed samples, \emph{progress credit assignment} for learning from verifier-confirmed progress, and \emph{structured feedback injection} for integrating feedback into the model's reasoning process.
Trained on sampled OpenR1-Math, MulFeRL outperforms supervised, self-distillation-based, and RLVR baselines in-domain and shows strong out-of-domain generalization. \footnote{We release our code at \url{https://anonymous.4open.science/r/MulFeRL}.}
\end{abstract}

\section{Introduction}
Reinforcement Learning with Verifiable Rewards (RLVR) has become a popular approach for improving reasoning capabilities in domains such as mathematics and code generation \citep{shao2024deepseekmath,cobbe2021trainingverifierssolvemath}. A verifier assigns each generated answer a scalar reward, e.g., correct or incorrect, which is used to optimize the policy \citep{deepseekai2025deepseekr1}. However, as model performance increases, training that relies solely on final outcome correctness yields diminishing gains and  eventually struggles to further improve the model \citep{zhang2025critique,yu2025dapo}.

We argue that this bottleneck stems from the sparsity and uninformative content of scalar rewards \citep{ye2025beyondcorrectness,wang2025grpo}. In early training stages or on harder problems, models often generate a large fraction of entirely incorrect solutions; under standard RLVR, failed solutions typically receive near-zero or negative rewards that merely indicate incorrectness, but provide little information about why the reasoning failed or how to improve it \citep{cai2025reinforcement,lightman2024let}. Consequently, the resulting learning signal suffers from poor credit assignment: gradient estimates are noisy, training becomes sample-inefficient, and progress slows or even plateaus, ultimately limiting final performance \citep{zheng2025survey}.

In contrast, when errors occur, humans typically leverage verbal feedback to localize mistakes and obtain fine-grained guidance for closing the gap to the target solution \citep{hattie2007power,kluger1996effects}. Inspired by this, we incorporate verbal feedback into model training: beyond indicating whether an answer is correct, natural language can identify key errors, missing steps, and actionable suggestions, providing a more informative, fine-grained learning signal for unsuccessful trajectories. This enables the model to learn from failed trajectories that would otherwise be treated as uniformly uninformative under outcome-only rewards.


Recent work has explored verbal feedback as richer supervision beyond sparse outcome-only rewards.
Directly optimizing verbal feedback as scalar rewards or feedback-revised supervision targets \citep{kim2024prometheus,liu2024chain} can compress its fine-grained diagnostics into coarse reward labels or imitation targets.
Self-distillation methods use feedback-conditioned or privileged policies as dense teachers \citep{hubotter2026reinforcement,zhao2026self}, but matching a privileged teacher distribution may over-compress uncertainty and self-correction traces that are important for robust reasoning \citep{kim2026does}.
RL-style feedback methods inject critiques or process guidance into policy optimization \citep{zhang2025critique,song2026expanding}, showing that feedback can improve refinement and exploration. However, they typically allocate feedback broadly across samples rather than targeting failed samples, where outcome-only rewards provide no usable contrast. Since samples with a successful trajectory are already outcome-informative, broad feedback allocation can under-exploit failed samples, where feedback is most needed.

Building on these observations, we focus on a central question: \textbf{how can verbal feedback be transformed into a stable learning signal that continues to drive RL training progress on failed samples?} Our key idea is that, during training, when a sample receives uniformly zero scalar reward and thus provides no informative learning signal, we introduce verbal feedback to guide regeneration and measure the resulting feedback-conditioned improvements, using them to construct an optimizable training objective, thereby converting verbal feedback into gradient signals.


We instantiate this idea in \textbf{MulFeRL}, a multi-turn, event-triggered RLVR framework with two stages: \emph{progress induction} and \emph{progress credit assignment}. In the progress induction stage, MulFeRL targets uniformly failed samples, using triggered feedback as a conditioning signal for iterative regeneration. This converts failure cases that originally lack reward contrast into verifier-certified progress. In the progress credit assignment stage, MulFeRL extracts learning signal from the contrast exposed by such progress. It uses GRPO to exploit within-group contrast after regeneration, and introduces Feedback-Contrastive Optimization (FCO) to recover cross-state signal from all-failed to all-successful transitions. MulFeRL further uses \emph{structured feedback injection} to integrate feedback into the model's reasoning process, improving its ability to use feedback during regeneration and learn from feedback-induced progress.

We train MulFeRL on a sampled subset of OpenR1-Math \citep{yan2026learning} and evaluate it on math benchmarks and OOD science/general reasoning tasks. MulFeRL consistently outperforms supervised, self-distillation-based, and RL-based baselines, with further analyses probing the roles of feedback-induced progress and credit assignment.

\section{Related Work}
\paragraph{Enhancing LLM Reasoning with Reinforcement Learning.} 
In verifiable domains such as mathematics and code, RLVR trains LLMs with rule-based outcome signals through online policy optimization, commonly using PPO-style methods and GRPO variants for efficient reasoning post-training \citep{schulman2017proximal,shao2024deepseekmath,liu2025understanding,yu2025dapo}. 
Recent work further extends RLVR-style training beyond single-turn reasoning to multi-turn and environment-interactive agents, incorporating environment feedback and turn-level credit assignment \citep{wang2025ragen,zeng2025reinforcing}. 
Despite these advances, RLVR remains constrained by sparse scalar rewards: failed samples receive little guidance on what went wrong or how to revise, often leading to training plateaus \citep{uesato2022solving,liu2026efficient}. 
We address this limitation by injecting verbal feedback during training, using it to guide regeneration and expose richer learning signals from failed samples.

\paragraph{Learning from Verbal Feedback.}
Recent work has explored verbal feedback as richer supervision beyond sparse outcome-only rewards.
Early methods convert feedback into scalar rewards or revised supervised targets \citep{kim2024prometheus,hancock2019learning,liu2024chain}, which eases optimization but collapses fine-grained diagnostics into coarse labels or imitation targets; self-distillation methods instead use feedback-conditioned or privileged policies as dense teachers \citep{hubotter2026reinforcement,zhao2026self}, but teacher-distribution matching may over-compress uncertainty and self-correction traces important for robust reasoning \citep{kim2026does}.
RL-style feedback methods inject critiques or process guidance into policy optimization \citep{zhang2025critique,song2026expanding}, but typically allocate feedback broadly rather than targeting uniformly failed RLVR samples, where outcome rewards provide no usable contrast.
Since samples with successful trajectories are already outcome-informative, such broad allocation can under-exploit the all-fail cases where feedback is most needed.
MulFeRL instead targets these cases and turns feedback-induced, verifier-certified progress into reward-grounded learning signals through multi-turn regeneration.

\section{Method}
\begin{figure*}[!t]
    \centering
    \includegraphics[width=\textwidth,height=6.5cm,keepaspectratio]{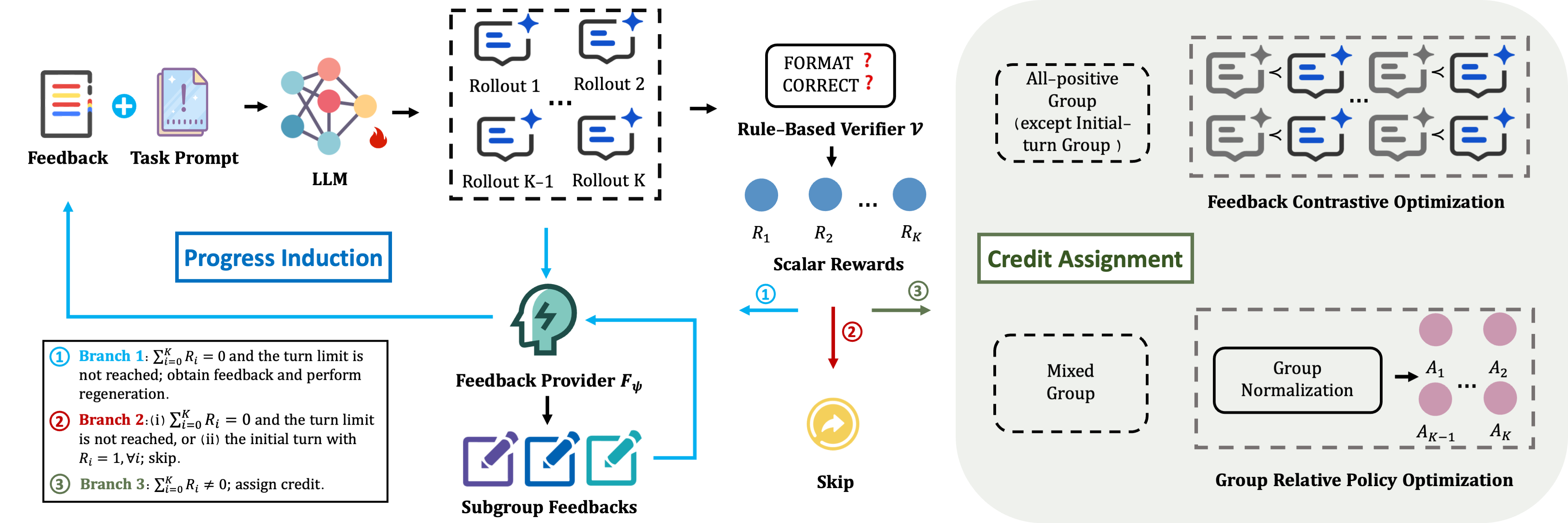}
    \caption{Illustration of MulFeRL.}
    \label{fig:main}
\end{figure*}

\label{sec:method}

\paragraph{Notation.}
Throughout this section, $x\sim\mathcal D$ denotes an input prompt, $\pi_\theta$ denotes the trainable policy, $\pi_{\mathrm{old}}$ denotes the frozen behavior policy for rollout collection, $\pi_{\mathrm{ref}}$ denotes the reference policy, and $\mathcal V$ denotes the verifier. For each outer update, we freeze the current policy as $\pi_{\mathrm{old}}$ and optimize $\pi_\theta$ with clipped importance ratios $\pi_\theta/\pi_{\mathrm{old}}$. At turn $t$, $\pi_{\mathrm{old}}$ samples a rollout group
$\mathcal G^{(t)}=\{y_i^{(t)}\}_{i=1}^{K}$ under context $c^{(t)}$, where $c^{(0)}=x$ and $y_i^{(t)}\sim\pi_{\mathrm{old}}(\cdot\mid c^{(t)})$. We write
$r_i^{(t)}=R_{\mathcal V}(x,y_i^{(t)})\in\{0,1\}$ for the verifier reward.
We denote the feedback provider by $F_\psi$, which produces feedback $f^{(t)}$ at turn $t$ and can be instantiated as a human user, a user-simulator model, or the policy model itself.

\subsection{Problem Formulation}
\label{sec:problem}

In GRPO-style RLVR, the frozen behavior policy collects $K$ rollouts for each prompt, and a verifier assigns binary rewards indicating whether each rollout is correct. GRPO updates rely on within-group relative advantages computed from these verifier rewards. A key difficulty arises for failed samples, where all $K$ rollouts fail verification.
In this case, the verifier provides no discrimination among candidates, causing the within-group advantage signal to collapse and providing no policy-gradient learning signal, which may lead to training plateaus.
Inspired by how humans use verbal feedback to localize errors and guide corrections, we introduce verbal feedback into RLVR to provide learning signal on failed samples, where the scalar verifier offers little discrimination. This raises a core question:
\textbf{How can we convert verbal feedback into a stable learning signal that continues to improve the policy on failed samples?}
Formally, our goal is to design a feedback-derived surrogate objective $\mathcal{J}_\theta$ that provides informative gradients and supports stable optimization:

\begin{equation}
\max_{\theta}\;
\mathbb{E}_{x\sim\mathcal{D}}\;
\mathbb{E}_{\mathcal{T}(x;\pi_{\mathrm{old}})}
\Big[\,\mathcal{J}_\theta\!\big(x,\mathcal{T}(x;\pi_{\mathrm{old}})\big)\,\Big],
\label{eq:surrogate_objective}
\end{equation}
where $\mathcal{T}(x;\pi_{\mathrm{old}})$ denotes the collected multi-turn feedback trajectory.

\subsection{Overview of MulFeRL}
\label{sec:overview}

To tackle the low-signal regime in RLVR, we propose \textbf{MulFeRL} (\textbf{Mul}ti-turn \textbf{Fe}edback-guided \textbf{R}einforcement \textbf{L}earning), an event-triggered RLVR framework that converts uniform-reward failures into verifier-certified progress events.

MulFeRL has two stages: \emph{progress induction} and \emph{progress credit assignment}. In the first stage, verbal feedback is used only as a conditioning intervention to help all-failed rollout groups escape the zero-reward state. In the second stage, the policy is updated only when the verifier exposes a usable discrimination signal. Thus, feedback changes the sampling condition, while learning remains gated by verifier-confirmed progress.

For a rollout group $\mathcal G^{(t)}=\{y_i^{(t)}\}_{i=1}^{K}$ under context $c^{(t)}$, with rewards $r_i^{(t)}=R_{\mathcal V}(x,y_i^{(t)})$, we define its reward state as:

\begin{equation}
z(\mathcal G^{(t)})
=
\begin{cases}
\textsc{Solved}, &
\sum_{i=1}^{K} r_i^{(t)}=K,
\\
\textsc{Contrastive}, &
0 < \sum_{i=1}^{K} r_i^{(t)} < K,
\\
\textsc{Failed}, &
\sum_{i=1}^{K} r_i^{(t)}=0.
\end{cases}
\label{eq:reward_state}
\end{equation}

The pipeline is organized around the location of verifier-exposed discrimination. 
For each prompt, MulFeRL first samples a feedback-free group under $c^{(0)}=x$ and verifies its reward state. If the group is \textsc{Contrastive}, the verifier already provides discrimination within the current state, and MulFeRL performs within-state credit assignment. If the group is \textsc{Solved}, no update is needed because the prompt is already solved without feedback. If the group is \textsc{Failed}, MulFeRL triggers feedback-guided regeneration until the verifier exposes progress or the turn budget is exhausted.

Once a failed group escapes the all-zero state, MulFeRL assigns credit according to where the new discrimination arises. A \textsc{Contrastive} regenerated group exposes within-state discrimination and is optimized with GRPO. A \textsc{Solved} regenerated group has no within-group contrast, but the adjacent \textsc{Failed}$\rightarrow$\textsc{Solved} transition exposes cross-state discrimination; MulFeRL therefore applies Feedback-Contrastive Optimization (FCO) to credit the transition. If all turns remain \textsc{Failed}, no update is applied.


\subsection{Event-Triggered Progress Induction}
\label{sec:progress_induction}

The key challenge in using verbal feedback for RL training is that feedback itself is not an optimizable learning signal. It is not a verifier reward and thus should not be optimized directly. Our key intuition is that effective verbal feedback should guide the model to turn an incorrect attempt into a correct one, so that the task improvements induced by feedback can be observed and used as an optimizable signal. Motivated by this intuition, the first stage of MulFeRL is \emph{progress induction}, which uses verbal feedback as a conditioning intervention to expose verifier-certified progress from failed rollout groups.
Accordingly, MulFeRL uses feedback only when the current rollout group is in the \textsc{Failed} state, and treats a transition from an all-failed condition to either a \textsc{Contrastive} or \textsc{Solved} group as induced progress. At a failed turn $t$, MulFeRL obtains feedback from the feedback provider:


\begin{equation}
f^{(t)}
\sim
F_\psi\!\left(
\cdot
\,\middle|\,
x,\;
\mathcal G^{(t)},\;
\{r_i^{(t)}\}_{i=1}^{K}
\right).
\label{eq:feedback}
\end{equation}
The feedback is a group-level, actionable summary of the current failures. It identifies the dominant failure mode and provides concrete guidance for revision\footnote{Since rollouts are typically long, we use a two-step aggregation: generate $g$ subgroup feedbacks and then summarize them into a single feedback (details in \S~\ref{app:alg}).}. We then inject the feedback into the next-turn context, $c^{(t+1)}=\operatorname{Inject}(x,f^{(t)})$, and regenerate $K$ rollouts with $y_i^{(t+1)}\sim\pi_{\mathrm{old}}(\cdot\mid c^{(t+1)})$.

The regenerated group is verified again. If a \textsc{Failed} group becomes \textsc{Contrastive} or \textsc{Solved}, MulFeRL treats it as induced progress and proceeds to credit assignment; otherwise, the loop continues until the turn budget is reached. By treating verbal feedback as an intervention measured by reward-state changes rather than as a direct optimization target, MulFeRL focuses extra sampling on low-signal prompts and converts feedback-induced progress into stable policy updates.

\subsection{Progress Credit Assignment}
\label{sec:credit_assignment}

Progress induction creates feedback-guided trajectories, but it does not by itself specify what should be credited. Since verbal feedback is used only as a conditioning intervention, MulFeRL treats credit assignment as an identifiability problem: a policy update is applied only when the verifier exposes a reward contrast that can be attributed to either candidates within the same state or progress across adjacent states. This view yields a unified principle: feedback may change the sampling condition, but credit is assigned only through verifier-certified discrimination.
This principle leads to two complementary objectives: GRPO for within-state discrimination, and FCO for cross-state discrimination that appears only across a \textsc{Failed}$\rightarrow$\textsc{Solved} transition.

\paragraph{Within-state credit assignment with GRPO.}
For a \textsc{Contrastive} group $\mathcal G^{(t)}$, verifier outcomes distinguish successful and failed rollouts under the same conditioning context. Since all candidates share the same prompt and feedback condition, relative rewards can be attributed to response quality rather than context changes. MulFeRL therefore applies the standard clipped GRPO objective:

\begin{equation}
\begin{aligned}
\mathcal L_{\mathrm{GRPO}}(x;\mathcal G^{(t)})
&=
-\frac{1}{K}\sum_{i=1}^{K}
\min\!\Big(
\rho_i^{(t)} A_i^{(t)},
\\
&\qquad\qquad
\operatorname{clip}(\rho_i^{(t)},1-\epsilon,1+\epsilon)
A_i^{(t)}
\Big)
\\
&\quad+
\beta_{\mathrm{KL}}
\mathrm{KL}\!\left(
\pi_\theta(\cdot\mid c^{(t)})
\middle\|
\pi_{\mathrm{ref}}(\cdot\mid c^{(t)})
\right).
\end{aligned}
\label{eq:grpo_obj}
\end{equation}
where $A_i^{(t)}$ is the group-normalized verifier reward and
$\rho_i^{(t)}=\pi_\theta(y_i^{(t)}\mid c^{(t)})/
\pi_{\mathrm{old}}(y_i^{(t)}\mid c^{(t)})$.
Thus, verifier-successful rollouts receive positive relative credit, while verifier-failed rollouts receive negative relative credit.



\paragraph{Cross-state credit assignment with FCO.}
However, feedback does not always expose within-state discrimination. In some cases, it is sufficiently effective that the regenerated group becomes \textsc{Solved}, i.e., all current rollouts are verifier-positive. In this case, all rollouts receive reward $1$, so within-group advantages collapse. The learning signal is not absent, but located across adjacent reward states: before the intervention, the same prompt produced only verifier failures; after the intervention, it produces only verifier successes. FCO is designed to credit this cross-state contrast rather than discard it as a reward-saturated group.


Let $\widehat{\mathcal Q}_{t}$ be $K$ sampled cross-state pairs $(y^+,y^-)$, where
$y^+$ comes from the current \textsc{Solved} group and $y^-$ from the immediately
preceding \textsc{Failed} group. We canonicalize rollouts by removing injected
feedback tokens from the likelihood objective, but not from the conditioning context
used for feedback-guided regeneration.
We randomly permute the failed rollouts when forming pairs, so
no same-index correspondence is assumed\footnote{The unbiasedness of this permutation estimator is proved in \S~\ref{app:theory:pair_sampling}.}.
Define the length-normalized log-ratio score $s_\theta(y;c)=|y|^{-1}\log\frac{\pi_\theta(y\mid c)}
{\pi_{\mathrm{ref}}(y\mid c)}$. The FCO loss is:

\begin{equation}
\begin{aligned}
\mathcal L_{\mathrm{FCO}}^{(t)}
&=
-\frac{1}{|\widehat{\mathcal Q}_{t}|}
\sum_{(y^+,y^-)\in\widehat{\mathcal Q}_{t}}
\log \sigma\!\Bigl(
\tau \bigl[
s_\theta(y^+;c^{(t)})
\\
&\qquad\qquad\qquad\qquad
- s_\theta(y^-;c^{(t-1)}) - \gamma
\bigr]
\Bigr).
\end{aligned}
\label{eq:fco_loss}
\end{equation}
where $\tau$ is an inverse temperature and $\gamma\ge0$ is an optional margin.



\paragraph{Unified routing.}
Let $t^\star=\min\{t:z(\mathcal G^{(t)})\neq\textsc{Failed}\}$ if the set is nonempty; otherwise no update is applied. For $\mathcal G^\star=\mathcal G^{(t^\star)}$ and $z^\star=z(\mathcal G^\star)$, MulFeRL uses
\begin{equation}
\mathcal L(x)
=
\begin{cases}
\mathcal L_{\mathrm{GRPO}}(x;\mathcal G^\star),
&
z^\star=\textsc{Contrastive},
\\[1mm]
\lambda\,\mathcal L_{\mathrm{FCO}}^{(t^\star)},
&
t^\star>0,\; z^\star=\textsc{Solved},
\\[1mm]
0,
&
\text{otherwise}.
\end{cases}
\label{eq:piecewise_loss}
\end{equation}
Thus, GRPO is used whenever the verifier exposes within-state contrast, while FCO is used only for feedback-induced \textsc{Failed}$\rightarrow$\textsc{Solved} transitions.

\begin{figure}[t]
\centering
\begin{tcolorbox}[
  title=Fixed feedback slot,
  colback=white,
  colframe=black!75,
  coltitle=white,
  colbacktitle=blue!45!black,
  fonttitle=\bfseries\scriptsize,
  boxrule=0.7pt,
  arc=1.2mm,
  width=\columnwidth,
  left=1.2mm,right=1.2mm,top=0.8mm,bottom=0.8mm,
]
\scriptsize\ttfamily
\textcolor{blue!55!black}{<thinking>}\\
\quad
\textcolor{red!55!black}{<feedback>}
\textcolor{red!55!black}{\{verbal feedback\}}
\textcolor{red!55!black}{</feedback>}\\
\quad \textcolor{gray!75!black}{\textit{rest of reasoning}}\\
\textcolor{blue!55!black}{</thinking>}
\ \textcolor{green!45!black}{solution}
\end{tcolorbox}
\caption{Fixed feedback slot used by MulFeRL.}
\label{fig:output_format}
\vskip -0.1in
\end{figure}

\subsection{Structured Feedback Injection}
\label{sec:injection}

Inspired by passage injection in retrieval-augmented generation~\citep{tang2025injecting},
MulFeRL injects verbal feedback into a fixed \texttt{<feedback>} slot in the
reasoning template, as shown in Figure~\ref{fig:output_format}, rather than
appending it to the prompt. The slot is empty at the feedback-free turn and is
filled with feedback from $F_\psi$ at regeneration turns. This fixed-slot design
treats feedback as an explicit conditioning intervention, embeds it more tightly
into the model's reasoning context, and keeps injected tokens identifiable for
masking and canonicalization.
During training, externally injected feedback tokens are masked out; a rollout is
assigned reward $1$ only when both answer correctness and schema compliance hold.
We prompt the feedback simulator to produce concise \emph{Issue} and
\emph{Fix steps} fields.\footnote{
\S~\ref{app:alg} provides algorithmic details, and
\S~\ref{app:theory} provides the theoretical analysis.
}



\section{Experiment}
\label{sec:experiments}

\begin{table*}[t]
\centering{
\begingroup
\footnotesize
\setlength{\tabcolsep}{3.3pt}
\renewcommand{\arraystretch}{1.08}
\begin{tabular}{@{}lccccccccc@{}}
\toprule
\multirow{2}{*}{\textbf{Method}} &
\multicolumn{5}{c}{\textbf{Math (ID)}} &
\multicolumn{3}{c}{\textbf{Science \& General (OOD)}} &
\multirow{2}{*}{\textbf{Avg.}} \\
\cmidrule(lr){2-6}\cmidrule(lr){7-9}
& \textbf{AMC23}
& \textbf{AIME24}
& \makecell[c]{\textbf{Olympiad}\\\textbf{Bench}}
& \makecell[c]{\textbf{MATH}\\\textbf{500}}
& \makecell[c]{\textbf{Minerva}\\\textbf{MATH}}
& \makecell[c]{\textbf{MMLU}\\\textbf{Pro}}
& \makecell[c]{\textbf{GPQA}\\\textbf{Diamond}}
& \makecell[c]{\textbf{Theorem}\\\textbf{QA}}
& \\
\midrule

\multicolumn{10}{@{}l}{\textit{Base Model}}\\
Qwen2.5-7B-Base
& 34.50 & 13.33 & 28.13 & 55.24 & 18.31 & 45.06 & 27.68 & 19.00 & 30.16 \\
\addlinespace[2pt]

\multicolumn{10}{@{}l}{\textit{Supervised Learning-based Finetuning}}\\
SFT
& 38.50 & 12.00 & 27.27 & 57.12 & 21.03 & 47.20 & 28.08 & 24.10 & 31.91 \\
RAFT
& 46.50 & 9.33 & 29.58 & 61.92 & 17.50 & 46.06 & 23.94 & 21.90 & 32.09 \\
CITL-FT
& 41.50 & 15.33 & 30.89 & 63.00 & 19.34 & 47.86 & 27.07 & 23.65 & 33.58 \\
\addlinespace[2pt]

\multicolumn{10}{@{}l}{\textit{Self-Distillation-based Finetuning}}\\
SDPO
& 41.50 & 14.67 & 35.91 & 69.72 & 28.24
& 50.90 & 32.53 & 37.20 & 38.83 \\
\addlinespace[2pt]

\multicolumn{10}{@{}l}{\textit{Reinforcement Learning-based Finetuning}}\\
GRPO
& 42.00 & 16.00 & 36.77 & 70.84 & 28.97 & 51.10 & 33.43 & 37.55 & 39.58 \\
Dr.GRPO
& 41.00 & 14.67 & 35.79 & 73.20 & 30.37 & 51.88 & 33.13 & 40.08 & 40.02 \\
Critique-GRPO
& 45.50 & 20.67 & 38.64 & 74.96 & 33.82 & 52.34 & 36.97 & 39.75 & 42.83 \\
RLTF-FM
& 42.00 & 15.33 & 36.05 & 69.88 & 28.46
& 50.76 & 32.93 & 37.65 & 39.13 \\
\addlinespace[2pt]

\multicolumn{10}{@{}l}{\textit{Multi-turn Feedback-guided Reinforcement Learning}}\\
MulFeRL
& \textbf{50.00} & \textbf{24.00} & \textbf{42.49} & \textbf{78.72} & \textbf{37.87}
& \textbf{54.89} & \textbf{38.99} & \textbf{43.08} & \textbf{46.26} \\

\midrule

\multicolumn{10}{@{}l}{\textit{Base Model}}\\
Qwen3-4B-Inst
& 59.00 & 46.00 & 45.13 & 76.68 & 42.43 & 58.56 & 35.05 & 39.85 & 50.34 \\
\addlinespace[2pt]

\multicolumn{10}{@{}l}{\textit{Supervised Learning-based Finetuning}}\\
SFT
& 62.50 & 47.33 & 49.17 & 78.12 & 46.32 & 60.46 & 36.26 & 41.93 & 52.76 \\
RAFT
& 62.00 & 46.00 & 48.34 & 78.68 & 45.29 & 59.53 & 37.47 & 41.05 & 52.30 \\
CITL-FT
& 62.00 & 48.00 & 48.55 & 79.64 & 47.13 & 60.18 & 36.87 & 42.65 & 53.13 \\
\addlinespace[2pt]

\multicolumn{10}{@{}l}{\textit{Self-Distillation-based Finetuning}}\\
SDPO
& 79.50 & 58.00 & 59.55 & 87.80 & 50.07
& 63.00 & 43.84 & 50.30 & 61.51 \\
\addlinespace[2pt]

\multicolumn{10}{@{}l}{\textit{Reinforcement Learning-based Finetuning}}\\
GRPO
& 78.50 & 57.33 & 59.05 & 87.84 & 51.25 & 62.18 & 43.94 & 49.85 & 61.24 \\
Dr.GRPO
& 79.00 & 56.67 & 60.36 & 88.28 & 49.19 & 63.33 & 42.83 & 50.68 & 61.29 \\
Critique-GRPO
& 84.00 & 62.67 & 62.40 & 88.64 & 50.29 & 64.67 & 45.45 & 51.05 & 63.65 \\
RLTF-FM
& 82.50 & 60.67 & 61.54 & 88.12 & 49.63
& 64.25 & 44.65 & 50.85 & 62.78 \\
\addlinespace[2pt]

\multicolumn{10}{@{}l}{\textit{Multi-turn Feedback-guided Reinforcement Learning}}\\
MulFeRL
& \textbf{90.50} & \textbf{68.00} & \textbf{68.13} & \textbf{90.24} & \textbf{55.96}
& \textbf{68.08} & \textbf{50.20} & \textbf{55.75} & \textbf{68.36} \\

\bottomrule
\end{tabular}
\endgroup}
\caption{Evaluation results (Pass@1) on mathematical reasoning (ID) and scientific/general reasoning (OOD) benchmarks. MulFeRL shows significant improvements across all datasets compared to RL baselines using paired t-test.}
\label{tab:main}
\end{table*}

\subsection{Experiment Settings}

\paragraph{Datasets.} 
For training, we randomly sample 4k instances from the reorganized 45k subset of OpenR1-Math-220k following \citet{yan2026learning}, without using any examples, labels, or solutions from the evaluation benchmarks. We evaluate on five math benchmarks (AMC 2023 \citep{numina_math_datasets}, AIME 2024 \citep{numina_math_datasets}, OlympiadBench \citep{he2024olympiadbenchchallengingbenchmarkpromoting}, MATH-500 \citep{hendrycks2021measuring} and Minerva-Math \citep{NEURIPS2022_18abbeef}), spanning routine to competition-level problems. For broader generalization, we additionally test on MMLU-Pro \citep{wang2024mmlu}, GPQA-Diamond \citep{rein2023gpqa} and TheoremQA \citep{chen2023theoremqa}. We report pass@1 averaged over 5 independent runs per dataset.\footnote{Dataset details and access links are provided in \S~\ref{app:datasets}.}

\paragraph{Baselines.}
We consider four groups of baselines:
(1) Base model: the backbone without any training;
(2) Supervised finetuning baselines: \textit{SFT} finetunes on the training set using supervised learning, \textit{RAFT} \citep{dong2023raft} finetunes on self-generated correct responses selected by rule-based verification, and \textit{CITL-FT} \citep{xi2024enhancingllmreasoningcritique} finetunes on self-generated and critique-refined responses;
(3) Self-distillation baselines: \textit{SDPO} \citep{hubotter2026reinforcement} distills feedback-conditioned rollouts from the current policy into the no-feedback policy;
(4) RL-based finetuning baselines: \textit{GRPO} \citep{deepseekai2025deepseekr1} optimizes binary outcome rewards; \textit{Dr.GRPO} \citep{liu2025understanding} and \textit{Critique-GRPO} \citep{zhang2025critique} are GRPO variants that respectively remove biased terms and use critiques for refinement; and \textit{RLTF-FM} \citep{song2026expanding} trains with multi-turn textual feedback and an auxiliary feedback-prediction loss, thereby internalizing feedback-driven improvements into the test-time first-turn policy.

\paragraph{Implementation Details.}
We use Qwen2.5-7B-Base \citep{DBLP:journals/corr/abs-2412-15115} and Qwen3-4B-Instruct-2507 \citep{qwen3technicalreport} (Qwen3-4B-Inst) as backbones, and GPT-4o \citep{openai_gpt4o_system_card_2024} as the feedback simulator $F_\psi$. For MulFeRL, we perform one initial rollout turn followed by up to two feedback-guided turns. In each turn, we sample 8 rollouts with temperature 1. Feedback text is capped at 1k tokens when used, and each generated solution is capped at 8k tokens during both training and evaluation. All baselines use the same sampling temperature and length limits unless stated otherwise. For all feedback-based methods, e.g., MulFeRL, SDPO, RLTF-FM, and Critique-GRPO, GPT-4o is used as the feedback provider. SFT methods are trained for 600 steps; SDPO and RL-style methods for 400 steps. For each method, we report the checkpoint with the best overall validation performance. At test time, we decode with temperature 1 and evaluate in a single-pass setting without external feedback unless specified otherwise. MulFeRL adopts the structured output format in Figure~\ref{fig:output_format} throughout training and evaluation; for other methods, we follow their default output formats.\footnote{A study on output-format is provided in \S~\ref{app:output_format}.}\footnote{More implementation details are provided in \S~\ref{app:hparams}.}

\subsection{Main Results}
\label{exp:main}
As shown in Table~\ref{tab:main}, MulFeRL achieves the best average performance under both backbones, with consistent gains across diverse in-domain math benchmarks. 
It also improves out-of-domain science and general reasoning tasks despite being trained primarily on math, indicating that MulFeRL exhibits cross-domain transfer rather than merely fitting the training domain. Overall, these results support our central design: MulFeRL targets low-signal failed groups, uses progress induction to convert feedback into optimizable learning signals, and integrates these signals into policy optimization through feedback-aware credit assignment, thereby improving reasoning and achieving robust gains across domains and backbones.

The comparison with feedback-based baselines further clarifies MulFeRL's advantages. We attribute part of the gain to its RL-based formulation: unlike distillation, MulFeRL optimizes the policy from feedback-derived signals rather than imitating feedback-refined outputs, which may better preserve the model's intrinsic reasoning behaviors, exploration patterns, and inductive biases. We also hypothesize that its focus on low-signal negative samples, together with multi-turn feedback and feedback-aware credit assignment, helps extract useful optimization signal from failed rollouts rather than merely exposing the model to feedback.

\begin{table}[t]
  \centering{
  \begingroup
  \footnotesize
  \setlength{\tabcolsep}{3.8pt}
  \renewcommand{\arraystretch}{1.10}
  \begin{tabular}{@{}lcc@{}}
    \toprule
    \textbf{Method} &
    \makecell[c]{\textbf{AIME24} ($\Delta$)} &
    \makecell[c]{\textbf{MATH500} ($\Delta$)} \\
    \midrule
    MulFeRL
    & \textbf{24.00} & \textbf{78.72} \\
    \midrule
    w/o Regeneration
    & 16.00\,(\textcolor{gray}{-8.00})
    & 70.84\,(\textcolor{gray}{-7.88}) \\
    w/o FCO
    & 19.67\,(\textcolor{gray}{-4.33})
    & 76.12\,(\textcolor{gray}{-2.60}) \\
    w/o Feedback Injection
    & 20.67\,(\textcolor{gray}{-3.33})
    & 77.68\,(\textcolor{gray}{-1.04}) \\
    \bottomrule
  \end{tabular}
  \endgroup}
  \caption{Ablation of MulFeRL on Qwen2.5-7B-Base.}
  \label{tab:ablation-MulFeRL-math}
  \vspace{-1.0em}
\end{table}

\subsection{Ablation Study}
As shown in Table~\ref{tab:ablation-MulFeRL-math}, we ablate three core components of MulFeRL: (1) feedback-guided regeneration, where no additional feedback turns are triggered and training reduces to single-turn GRPO; (2) FCO, where regeneration is retained but cross-state credit assignment is removed, leaving only GRPO updates on contrastive groups; and (3) feedback injection, where the fixed-slot structured injection is replaced with a plain prompting variant that provides the same feedback content without the injection structure.
All ablations degrade performance, showing that each component is useful. Removing feedback-guided regeneration causes the largest degradation, confirming that progress induction is crucial for recovering learning signal from all-failed rollout groups. Removing FCO substantially hurts performance, showing that regeneration alone does not fully convert feedback-induced progress into trainable signal: when feedback moves an all-failed group to an all-solved group, within-state GRPO has no reward contrast to optimize, and cross-state FCO is needed to consolidate the improvement. Finally, replacing structured feedback injection with plain prompting leads to consistent declines, suggesting that the fixed-slot injection helps the model condition on feedback more reliably. Overall, MulFeRL benefits from combining feedback-guided progress induction, cross-state credit assignment, and structured feedback utilization.

\begin{table}[t]
  \centering
  {
  \footnotesize
  \setlength{\tabcolsep}{3.2pt}
  \renewcommand{\arraystretch}{1.08}
  \begin{tabular}{@{}lcccc@{}}
    \toprule
    \textbf{Simulator} &
    \textbf{SDPO} &
    \textbf{RLTF} &
    \makecell[c]{\textbf{Critique}\\\textbf{GRPO}} &
    \textbf{MulFeRL} \\
    \midrule
    Qwen3-1.7B (Thinking) & 57.86 & 58.42 & 58.18 & 61.32 \\
    Qwen3-4B-Inst (Self) & 60.06 & 59.74 & 61.35 & 65.62 \\
    GPT-4o-mini          & 60.84 & 62.36 & 62.12 & 66.84 \\
    GPT-4o               & 61.51 & 62.78 & 63.65 & \textbf{68.36} \\
    \bottomrule
  \end{tabular}
  }
  \caption{Average Pass@1 across all benchmarks when training feedback-based methods with different feedback simulators on Qwen3-4B-Inst. The \textit{self} simulator uses the frozen base Qwen3-4B-Inst checkpoint without additional simulator training.}
    \label{tab:simulator_qwen3}

  \vspace{-0.4em}
  
\end{table}

\begin{figure}[t]
  \begin{center}
    \centerline{\includegraphics[width=\columnwidth]{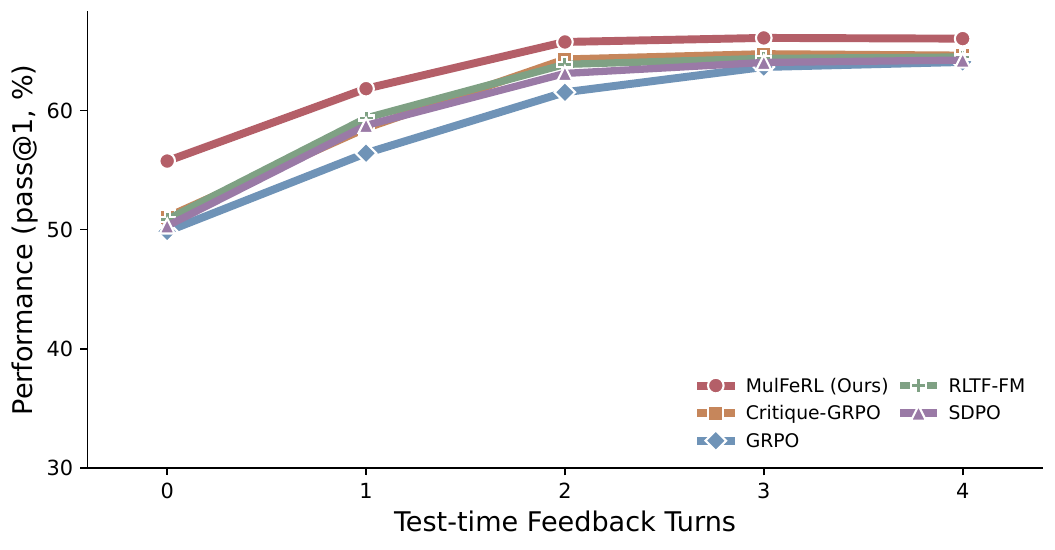}}
    \caption{Test-time multi-turn feedback results on TheoremQA (Qwen3-4B-Inst).}
    \label{fig:testtime_multiturn}
  \end{center}
  \vspace{-2.0em}
\end{figure}
\section{Analysis}

\subsection{Impact of Feedback Simulators}
\label{simulation}

Table~\ref{tab:simulator_qwen3} compares MulFeRL trained with feedback from different simulators. We observe a consistent trend that higher-quality simulators lead to better downstream performance. Using the base backbone itself as the feedback simulator already yields a strong MulFeRL model (Table~\ref{tab:simulators_full_qwen3} for full results). This suggests that even without guidance from a stronger external simulator, MulFeRL can improve reasoning via self-feedback. Moreover, stronger simulators provide larger gains, with GPT-4o-mini improving over self-feedback and GPT-4o further boosting performance. We hypothesize that stronger simulators yield more informative feedback, enabling better regeneration and denser learning signals that improve generalization on challenging math reasoning tasks.

\subsection{Test-time Multi-turn Feedback}
\label{inference_time}
We evaluate test-time multi-turn feedback on TheoremQA to assess whether MulFeRL training improves a model’s ability to use external natural-language feedback for inference-time revision. GPT-4o provides feedback using the same prompts as in training, without revealing answers. Figure~\ref{fig:testtime_multiturn} shows that more feedback turns improve all methods, while MulFeRL consistently achieves the best performance across turns. This indicates that MulFeRL can exploit external feedback effectively to refine its solutions and improve performance under multi-turn feedback. Performance also tends to level off in later turns, suggesting diminishing returns as the feedback simulator becomes the dominant limiting factor.

\subsection{Training efficiency}
\begin{figure}[t]
    \centering
    \begin{subfigure}[t]{\columnwidth}
        \centering
        \includegraphics[width=\linewidth]{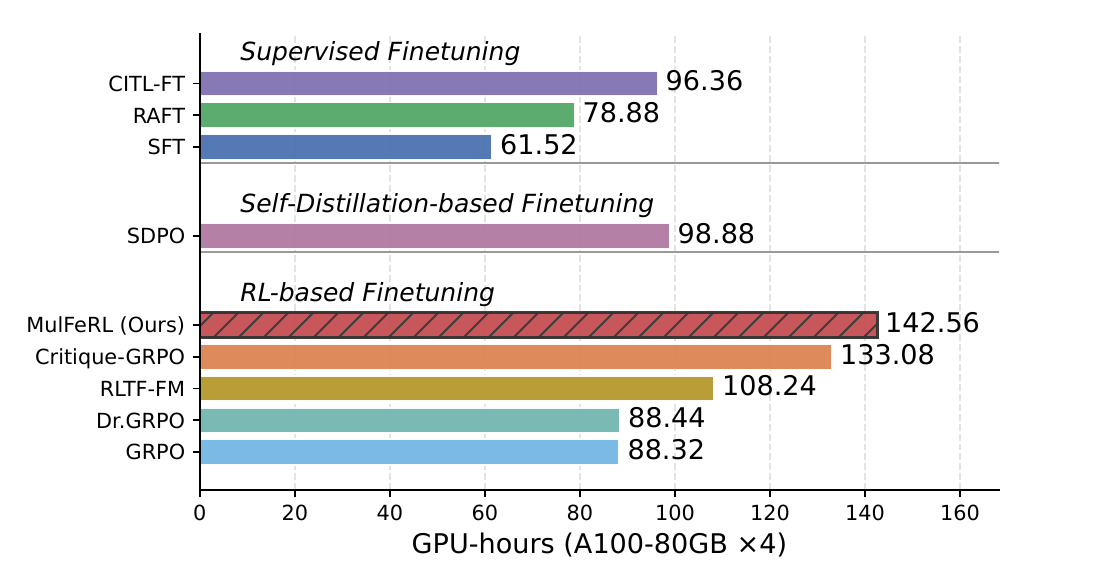}
        \caption{Training cost (GPU-hours).}
        \label{fig:efficiency-gpu-hours}
    \end{subfigure}
    \begin{subfigure}[t]{\columnwidth}
        \centering
        \includegraphics[width=\linewidth]{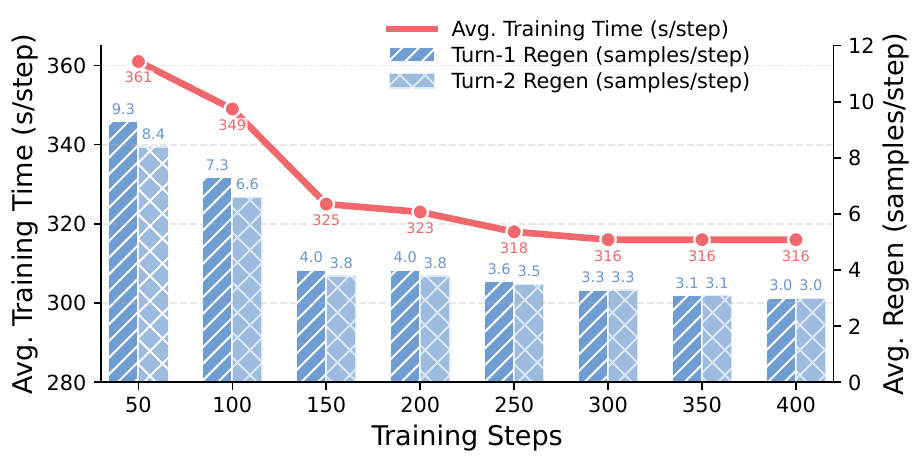}
        \caption{MulFeRL training efficiency across steps.}
        \label{fig:efficiency-step-time}
    \end{subfigure}
    \caption{Training efficiency (Qwen3-4B-Inst). (a) Training cost across methods. (b) MulFeRL efficiency across training steps: average training time per step and the average number of regenerated samples per step in Turn-1/Turn-2, reported as averages over 50-step intervals.}
    \label{fig:efficiency}
    \vskip -0.1in
\end{figure}
We measure training cost under the unified setup described in the implementation details. As shown in Figure~\ref{fig:efficiency-gpu-hours}, MulFeRL incurs higher training cost than RL baselines, primarily due to its multi-turn regeneration design. When sampled candidates fail, MulFeRL performs feedback-guided regeneration, introducing additional decoding and verification overhead. Figure~\ref{fig:efficiency-step-time} further illustrates this effect: regeneration increases the per-step time, with the overhead being most pronounced early in training when failures are more frequent and regeneration is triggered more often. 
Despite the higher per-step cost, MulFeRL can reach strong performance in fewer optimization steps because feedback-guided regeneration converts low-signal failed groups into verifier-certified progress, while the credit-assignment objective exploits identifiable discrimination, making each update more sample-efficient (see \S~\ref{app:extra:convergence} for details).
Thus in practice, overall cost can also be reduced by using a smaller step budget (e.g., early stopping) and optionally annealing regeneration as the policy improves, which mitigates per-step overhead while retaining substantial gains.

\subsection{Impact of Multi-turn Training}
\begin{figure}[t]
  \begin{center}
    \centerline{\includegraphics[width=\columnwidth]{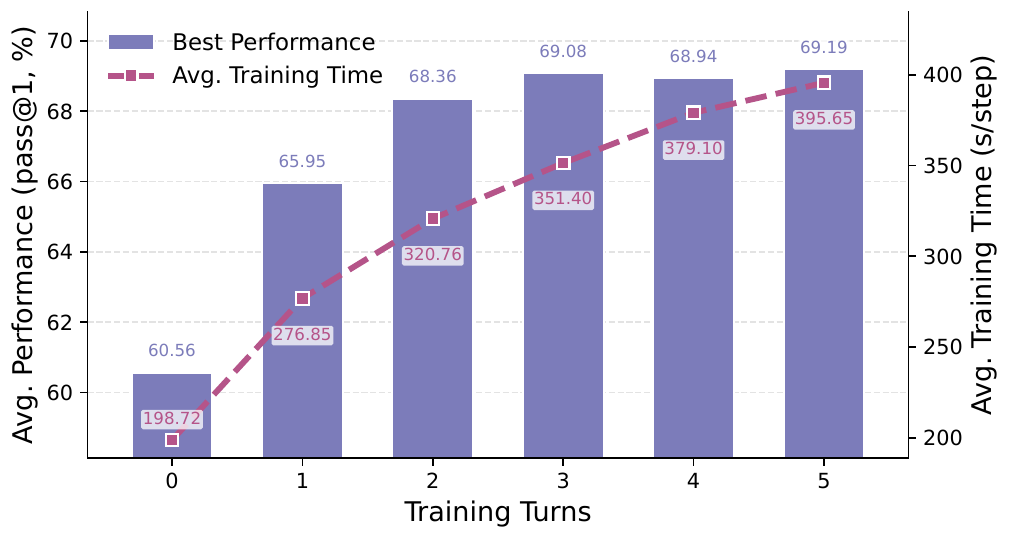}}
\caption{Impact of training turns on effectiveness and efficiency (Qwen3-4B-Inst).}
    \label{fig:training_turns}
  \end{center}
  \vspace{-2.0em}
\end{figure}
Figure~\ref{fig:training_turns} shows how the training-time per-step regeneration turn budget affects both effectiveness and efficiency. Increasing the number of turns yields rapid initial gains, suggesting that feedback-conditioned regeneration provides stronger learning signal early on. The improvements then saturate, with diminishing returns beyond a small number of turns. In contrast, the average training time per step increases steadily as more turns introduce additional decoding and verification overhead. Overall, the figure highlights a performance--cost trade-off and motivates using a small-to-moderate turn budget that captures most of the benefit while keeping training cost manageable.

\section{Conclusion}
Outcome-only scalar rewards in RLVR are often sparse on failed samples, leaving all-failed rollout groups with little usable signal and causing training to plateau. We propose MulFeRL, a multi-turn feedback-guided RLVR framework that targets these low-signal groups, injects verbal feedback through a structured format, uses progress induction to turn feedback-driven revisions into optimizable learning signals, and incorporates them into policy optimization through feedback-aware credit assignment. Experiments across two backbones show consistent gains over strong feedback-based baselines, with further analyses revealing the underlying mechanisms.

\section*{Limitations}

MulFeRL incurs additional training overhead compared with single-turn RLVR baselines, since multi-turn feedback-guided regeneration introduces extra decoding and verification. This cost is especially noticeable when failures are frequent, and can further increase when a strong external model is used as the feedback simulator. Nevertheless, our experiments show that MulFeRL remains effective even with weaker or self-generated feedback, suggesting that cheaper simulators, early stopping, and annealed regeneration can help reduce practical cost.

MulFeRL also inherits common limitations of learning from verbal feedback. Its effectiveness depends on reliable outcome verification and useful feedback: stronger simulators may improve results but add cost and reproducibility constraints, while weaker or noisy feedback may limit the recoverable signal. Our appendix experiments indicate that MulFeRL is relatively robust across feedback sources, but developing cheaper and more reliable feedback mechanisms remains an important future direction.

In addition, the current design uses rollout-level feedback and a conservative triggering strategy: feedback is applied only to all-failed groups, and cross-turn preference pairs are constructed with a simple pairing rule. While this targets the low-signal regime where standard RLVR struggles, it may miss finer-grained improvements in partially correct or near-miss rollouts. Future work can extend MulFeRL to step-level or process-level feedback, adaptive triggering, and richer credit-assignment schemes.

\bibliography{main}
\appendix
\appendix
\section{Algorithm Details}
\label{app:alg}

This appendix provides the full algorithmic specification of MulFeRL, including
notation, reward-state routing, feedback aggregation, structured injection,
token masking, canonicalization, and FCO pair sampling.

\subsection{Components and Notation}
\label{app:alg:notation}

\paragraph{Policy, behavior policy, and reference policy.}
Let $\pi_\theta$ denote the trainable policy, $\pi_{\mathrm{old}}$ the frozen
behavior policy used to sample rollouts for the current outer update, and
$\pi_{\mathrm{ref}}$ a fixed reference policy. At the beginning of each outer
update, we freeze the current policy as $\pi_{\mathrm{old}}$. All rollouts used
for this update are sampled from $\pi_{\mathrm{old}}$, while the loss is
optimized with respect to $\pi_\theta$. This follows the standard PPO-style
on-policy protocol, where trajectories are sampled from a behavior policy and
then optimized using a clipped surrogate objective~\citep{schulman2017proximal}.
GRPO follows the same policy-optimization convention while replacing a learned
critic with group-relative reward normalization~\citep{shao2024deepseekmath}.
The reference policy $\pi_{\mathrm{ref}}$ is used for the KL regularization term
in the GRPO branch and for the reference-normalized score in the FCO branch.

\paragraph{Prompt, turns, and rollout groups.}
For each prompt $x\sim\mathcal D$, MulFeRL runs a turn-based feedback trajectory
for at most $T$ turns. At turn $t$, the policy samples a group of $K$ rollouts:

\[
\mathcal G^{(t)}
=
[y_1^{(t)},\ldots,y_K^{(t)}],
\qquad
y_i^{(t)}\sim \pi_{\mathrm{old}}(\cdot\mid c^{(t)}),
\label{eq:app_group_sampling}
\]
where $c^{(t)}$ is the turn-specific conditioning context. The square brackets
indicate an ordered list rather than a mathematical set; duplicate rollouts are
preserved.

The initial context is feedback-free:

\[
c^{(0)}=x.
\]
For $t\ge 1$, the context is obtained by injecting the previous-turn feedback
into a fixed feedback slot:

\[
c^{(t)}
=
\operatorname{Inject}(x,f^{(t-1)}).
\]

\paragraph{Verifier and reward state.}
The verifier $\mathcal V$ assigns a binary reward:

\[
r_i^{(t)}
=
R_{\mathcal V}(x,y_i^{(t)})
\in\{0,1\}.
\]
The reward state of a group is:

\begin{equation}
z(\mathcal G^{(t)})
=
\begin{cases}
\textsc{Solved}, &
\sum_{i=1}^{K} r_i^{(t)}=K,
\\
\textsc{Contrastive}, &
0 < \sum_{i=1}^{K} r_i^{(t)} < K,
\\
\textsc{Failed}, &
\sum_{i=1}^{K} r_i^{(t)}=0.
\end{cases}
\label{eq:app_reward_state}
\end{equation}
A \textsc{Contrastive} group is optimized with GRPO. A feedback-induced
\textsc{Solved} group following a preceding \textsc{Failed} group is optimized
with FCO. A \textsc{Failed} group triggers feedback-guided regeneration unless
the turn budget has been exhausted.

\paragraph{Feedback provider.}
The feedback provider $F_\psi$ can be instantiated as a human annotator, an
external LLM, or the policy model itself via self-feedback. At a failed turn
$t$, it produces group-level feedback:

\[
f^{(t)}
\sim
F_\psi
\left(
\cdot
\,\middle|\,
x,\mathcal G^{(t)},\{r_i^{(t)}\}_{i=1}^{K}
\right).
\]
The feedback is used only as a conditioning intervention for the next turn. It
is not treated as a reward label and is not directly optimized. If the policy
model itself is used as the feedback provider, the generated feedback text is
treated as stop-gradient conditioning text.

\subsection{Reward Design and Verifier Criteria}
\label{app:alg:reward}

We use a binary verifiable reward:

\begin{equation}
R_{\mathcal V}(x,y)
=
\begin{cases}
1, & \substack{\text{if $y$ is answer-correct}\\
\text{and schema-compliant}},\\
0, & \text{otherwise}.
\end{cases}
\label{eq:app_reward_design}
\end{equation}

\paragraph{Answer correctness.}
A response is answer-correct if its final answer satisfies the task-specific
verification rule implemented by $\mathcal V$.

\paragraph{Schema compliance.}
A response is schema-compliant if it follows the structured output template,
including the required delimiters and final-answer format. Schema compliance is
included only to make automatic parsing and verification well-defined. If either
answer correctness or schema compliance fails, the verifier conservatively
returns zero reward.

\subsection{Multi-turn Feedback-guided Regeneration Loop}
\label{app:alg:loop}

MulFeRL applies feedback only when the current group is \textsc{Failed}. For
each prompt, the algorithm optimizes the first turn that provides an informative
update signal and then terminates the prompt. This prevents a single prompt from
contributing multiple highly correlated updates and avoids over-weighting long
feedback trajectories.

At turn $t$, after sampling and verification, the routing rule is:

\begin{itemize}[leftmargin=*, itemsep=1pt, topsep=2pt]
    \item \textbf{\textsc{Contrastive}.}
    If $\mathcal G^{(t)}$ contains both successes and failures, MulFeRL applies
    the GRPO loss on $\mathcal G^{(t)}$ and terminates this prompt. This case
    provides within-context success--failure contrast.

    \item \textbf{\textsc{Solved}.}
    If $t=0$ and $\mathcal G^{(0)}$ is already \textsc{Solved}, MulFeRL skips
    optimization because the prompt is solved without feedback and provides no
    within-group contrast. If $t>0$ and $\mathcal G^{(t)}$ is \textsc{Solved},
    then the immediately preceding group $\mathcal G^{(t-1)}$ must be
    \textsc{Failed}, because feedback is triggered only on failed groups.
    MulFeRL applies FCO using $\mathcal G^{(t-1)}$ as the all-failed anchor and
    $\mathcal G^{(t)}$ as the all-positive regenerated group.

    \item \textbf{\textsc{Failed}.}
    If $\mathcal G^{(t)}$ is \textsc{Failed} and $t<T-1$, MulFeRL queries the
    feedback provider, injects the feedback into the next-turn context, and
    regenerates $\mathcal G^{(t+1)}$. If $t=T-1$, no progress has been observed
    within the turn budget, and the prompt receives zero loss.
\end{itemize}

\paragraph{First informative turn.}
Define:

\[
\begin{aligned}
t^\star
=
\min\bigl\{
t:\;&z(\mathcal G^{(t)})=\textsc{Contrastive}
\\
&\text{or }
\bigl(t>0 \wedge z(\mathcal G^{(t)})=\textsc{Solved}\bigr)
\bigr\},
\end{aligned}
\]
if such a turn exists. MulFeRL updates only at $t^\star$. If
$z(\mathcal G^{(t^\star)})=\textsc{Contrastive}$, it uses GRPO. If
$t^\star>0$ and $z(\mathcal G^{(t^\star)})=\textsc{Solved}$, it uses FCO
anchored by $\mathcal G^{(t^\star-1)}$. If no such $t^\star$ exists, the prompt
receives zero loss. An initial \textsc{Solved} group at $t=0$ is skipped rather
than treated as informative because it provides neither a failure anchor nor
within-group reward variation.

\subsection{Feedback Aggregation for Long Rollouts}
\label{app:alg:feedback_agg}

Reasoning rollouts can be long, making it difficult for a feedback provider to
produce concise and reliable feedback over all candidates at once. We therefore
use a two-stage feedback aggregation procedure.

\paragraph{Stage 1: subgroup feedback.}
Given an all-failed group:

\[
\mathcal G^{(t)}
=
[y_1^{(t)},\ldots,y_K^{(t)}],
\]
we randomly partition it into $g$ subgroups:

\[
\mathcal S_1^{(t)},\ldots,\mathcal S_g^{(t)}.
\]
For each subgroup, we query the feedback provider:

\begin{equation}
\begin{aligned}
f_j^{(t)}
&\sim
F_\psi\!\left(
\cdot
\,\middle|\,
\begin{gathered}
x,\\
[y_i^{(t)}:i\in\mathcal S_j^{(t)}],\\
[r_i^{(t)}:i\in\mathcal S_j^{(t)}]
\end{gathered}
\right),
\\
&\hspace{2em} j=1,\ldots,g.
\end{aligned}
\label{eq:app_subgroup_feedback}
\end{equation}
Each subgroup feedback identifies common failure patterns and proposes
actionable revision steps.

\paragraph{Stage 2: group-level summarization.}
The subgroup feedbacks are summarized into a single group-level feedback:

\begin{equation}
f^{(t)}
\sim
\operatorname{Summarize}
\left(
\cdot
\,\middle|\,
x,\,[f_1^{(t)},\ldots,f_g^{(t)}]
\right).
\label{eq:app_feedback_summarize}
\end{equation}
The final feedback follows the structured format used for injection:

\[
\begin{aligned}
\textit{Issue:}\;&\text{root mistake or dominant failure mode},\\
\textit{Fix steps:}\;&\text{actionable revision plan}.
\end{aligned}
\]
When rollouts are short, we set $g=1$ and skip the subgroup aggregation step.
The values of $K$, $g$, and $T$ are reported in Appendix~\ref{app:hparams}.

\subsection{Structured Feedback Injection and Token Masking}
\label{app:alg:inject_mask}

MulFeRL injects feedback into a fixed \texttt{<feedback>} slot inside the
reasoning template. This design is motivated by prior work showing that
language feedback can improve subsequent attempts through iterative refinement
or verbal reflection~\citep{madaan2023self,shinn2023reflexion}. It is
also related in interface design to passage injection, which explicitly places
external information inside the reasoning process rather than merely appending
it to the input~\citep{tang2025injecting}. In MulFeRL, the injected content is
feedback rather than retrieved evidence.

\paragraph{Injection interface.}
In the default setting, the initial feedback slot is empty:

\[
c^{(0)}=x.
\]
At regeneration turns, externally produced feedback $f^{(t-1)}$ is inserted
into the slot as Figure~\ref{fig:output_format} shows.
The model then continues generation conditioned on the injected feedback. If a
self-feedback variant is used, it is treated as a separate variant; the default
algorithm and the theoretical analysis assume that the initial context is
feedback-free.

\paragraph{Token masking.}
Externally injected feedback tokens are treated as conditioning tokens and are
masked out from the optimization loss. Let $M^{(t)}(y)$ denote the set of token
indices in rollout $y$ that correspond to model-generated reasoning and answer
tokens at turn $t$. Injected feedback tokens are excluded from $M^{(t)}(y)$.
Define the canonicalized body:

\[
b^{(t)}(y)
=
(y_\ell)_{\ell\in M^{(t)}(y)}.
\]
The masked conditional log-likelihood is:

\begin{equation}
\log \pi_\theta(b^{(t)}(y)\mid c^{(t)})
=
\sum_{\ell\in M^{(t)}(y)}
\log \pi_\theta
\left(
y_\ell
\,\middle|\,
c^{(t)},y_{<\ell}
\right).
\label{eq:app_masked_logprob}
\end{equation}
The same masking convention is used for $\pi_{\mathrm{old}}$ and
$\pi_{\mathrm{ref}}$.

Importantly, canonicalization removes injected feedback from the likelihood
objective, but it does not remove feedback from the conditioning context. For
$t>0$, the probability in Eq.~\eqref{eq:app_masked_logprob} is still computed
under the original feedback-augmented context $c^{(t)}$.

\subsection{Loss Interfaces}
\label{app:alg:losses}

\paragraph{GRPO branch.}
For a \textsc{Contrastive} group $\mathcal G^{(t)}$, MulFeRL computes
group-relative advantages:

\[
\begin{aligned}
\bar r^{(t)}
&=
\frac{1}{K}\sum_{i=1}^{K}r_i^{(t)},\\
\hat\sigma_r^{(t)}
&=
\sqrt{
\frac{1}{K}\sum_{i=1}^{K}
\left(r_i^{(t)}-\bar r^{(t)}\right)^2
},\\
\hat A_i^{(t)}
&=
\frac{
r_i^{(t)}-\bar r^{(t)}
}{
\hat\sigma_r^{(t)}+\epsilon_{\mathrm{adv}}
}.
\end{aligned}
\]
For compactness, write $M_i^{(t)}=M^{(t)}(y_i^{(t)})$. For an unmasked token
$\ell\in M_i^{(t)}$, define the PPO-style importance ratio:

\[
\rho_{i,\ell}^{(t)}(\theta)
=
\frac{
\pi_\theta
\left(
y_{i,\ell}^{(t)}
\,\middle|\,
c^{(t)},y_{i,<\ell}^{(t)}
\right)
}{
\pi_{\mathrm{old}}
\left(
y_{i,\ell}^{(t)}
\,\middle|\,
c^{(t)},y_{i,<\ell}^{(t)}
\right)
}.
\]
The GRPO branch uses the clipped group-relative policy loss:

\begin{align}
\mathcal L_{\mathrm{GRPO}}^{(t)}
&=
-\frac{1}{K}
\sum_{i=1}^{K}
\frac{1}{|M_i^{(t)}|}
\sum_{\ell\in M_i^{(t)}}
\min\Bigl(
\rho_{i,\ell}^{(t)}(\theta)\hat A_i^{(t)},
\nonumber\\
&\quad
\operatorname{clip}
\bigl(
\rho_{i,\ell}^{(t)}(\theta),
1-\epsilon_{\mathrm{clip}},
1+\epsilon_{\mathrm{clip}}
\bigr)
\hat A_i^{(t)}
\Bigr)
\nonumber\\
&\quad
+\beta_{\mathrm{KL}}
\mathcal R_{\mathrm{KL}}^{(t)}.
\label{eq:app_grpo_loss}
\end{align}
Here $\mathcal R_{\mathrm{KL}}^{(t)}$ denotes the implementation's
reference-policy KL regularizer computed over the same unmasked generated
tokens. The exact KL estimator follows the implementation used in the main
experiments.

\paragraph{FCO branch.}
FCO is used only for a \textsc{Failed}$\rightarrow$\textsc{Solved} transition.
For a canonicalized response body generated under context $c$, define the
length-normalized reference score:

\begin{equation}
s_\theta(c,y)
=
\frac{1}{|b(y)|}
\left[
\log\pi_\theta(b(y)\mid c)
-
\log\pi_{\mathrm{ref}}(b(y)\mid c)
\right],
\label{eq:app_fco_score}
\end{equation}
where the log-likelihoods are computed using the masking convention in
Eq.~\eqref{eq:app_masked_logprob}. This reference-normalized form is analogous
in spirit to pairwise preference objectives such as DPO, but here the pairwise
ordering is produced by an automatic verifier and an adjacent
\textsc{Failed}$\rightarrow$\textsc{Solved} transition rather than by human
preference annotations~\citep{rafailov2023direct}.

For a positive--negative pair $(y^+,y^-)$ with contexts $(c^+,c^-)$, define:

\[
D_\theta(y^+,y^-)
=
s_\theta(c^+,y^+)
-
s_\theta(c^-,y^-).
\]
The pairwise FCO loss is:

\begin{equation}
\ell_{\mathrm{FCO}}(y^+,y^-)
=
-\log
\sigma
\left(
\tau
\left[
D_\theta(y^+,y^-)-\gamma
\right]
\right),
\label{eq:app_fco_pair_loss}
\end{equation}
where $\tau>0$ is an inverse-temperature parameter and $\gamma\ge 0$ is a
margin. Given a pair set $\mathcal Q_{\mathrm{FCO}}^{(t)}$, the FCO branch uses:

\begin{equation}
\mathcal L_{\mathrm{FCO}}^{(t)}
=
\frac{1}{|\mathcal Q_{\mathrm{FCO}}^{(t)}|}
\sum_{(y^+,y^-)\in\mathcal Q_{\mathrm{FCO}}^{(t)}}
\ell_{\mathrm{FCO}}(y^+,y^-).
\label{eq:app_fco_loss}
\end{equation}

\subsection{Canonicalization for FCO}
\label{app:alg:canonicalization}

FCO compares responses generated under different conditioning contexts. To
ensure that this comparison is made over model-generated content rather than
externally supplied feedback, we use the canonicalized body $b^{(t)}(y)$ defined
in Section~\ref{app:alg:inject_mask}. The canonicalized body determines which
tokens enter the likelihood objective. The turn-specific context remains
unchanged.

This distinction is important because FCO compares the current all-positive
group $\mathcal G^{(t)}$ with the immediately preceding all-failed group
$\mathcal G^{(t-1)}$. These groups may have different feedback contexts, so the
objective should compare generated solutions conditioned on their respective
contexts, not the injected feedback text itself.

\subsection{FCO Pair Construction and Sampling}
\label{app:alg:fco_pairs}

FCO is used only for a \textsc{Failed}$\rightarrow$\textsc{Solved} transition.
Let $\mathcal G^{(t-1)}$ be the preceding all-failed group under context
$c^{(t-1)}$, and let $\mathcal G^{(t)}$ be the current all-positive group under
context $c^{(t)}$.

We construct ordered lists of positive and negative examples:

\[
\begin{aligned}
\mathcal Y_{+}^{(t)}
&=
\bigl[(c^{(t)},b^{(t)}(y_i^{(t)}))\bigr]_{i=1}^{K},
\\
\mathcal Y_{-}^{(t-1)}
&=
\bigl[(c^{(t-1)},b^{(t-1)}(y_i^{(t-1)}))\bigr]_{i=1}^{K}.
\end{aligned}
\]
These are ordered multisets rather than mathematical sets. Duplicate
canonicalized responses are preserved, so both lists have length $K$ even if two
rollouts yield identical bodies after canonicalization.

The full FCO pair set is the all-pair Cartesian product:

\begin{equation}
\mathcal Q_{\mathrm{FCO,full}}^{(t)}
=
\mathcal Y_{+}^{(t)}
\times
\mathcal Y_{-}^{(t-1)}.
\label{eq:app_fco_full_pairs}
\end{equation}

\paragraph{Permutation-based sampling.}
Using all $K^2$ pairs may be expensive. By default, we use a permutation-based
estimator. Write:

\[
\begin{aligned}
\mathcal Y_{+}^{(t)}
&=
[y_1^+,\ldots,y_K^+],
\\
\mathcal Y_{-}^{(t-1)}
&=
[y_1^-,\ldots,y_K^-],
\end{aligned}
\]
where each $y_j^+$ or $y_i^-$ denotes a context-body pair. We sample a random
permutation $\pi$ of $\{1,\ldots,K\}$ and construct:

\begin{equation}
\widehat{\mathcal Q}_{\mathrm{FCO}}^{(t)}
=
[
(y_j^+,y_{\pi(j)}^-):j=1,\ldots,K
].
\label{eq:app_fco_sampled_pairs}
\end{equation}
This avoids assuming any same-index correspondence between turns while using
each rollout once per update.

For any pairwise loss $\ell(y^+,y^-)$, the permutation estimator is unbiased for
the full all-pair average:

\begin{equation}
\mathbb E_{\pi}
\left[
\frac{1}{K}
\sum_{j=1}^{K}
\ell(y_j^+,y_{\pi(j)}^-)
\right]
=
\frac{1}{K^2}
\sum_{j=1}^{K}
\sum_{i=1}^{K}
\ell(y_j^+,y_i^-).
\label{eq:app_perm_unbiased}
\end{equation}
For notational simplicity, the main text writes
$\mathcal Q_{\mathrm{FCO}}^{(t)}$ to denote either the full pair set
$\mathcal Q_{\mathrm{FCO,full}}^{(t)}$ or its sampled estimator
$\widehat{\mathcal Q}_{\mathrm{FCO}}^{(t)}$.

\subsection{Verifier Gating and Reference Anchoring}
\label{app:alg:gating}

MulFeRL does not directly optimize feedback text. Feedback only changes the
next-turn conditioning context. A policy update is applied only when the
verifier observes either a \textsc{Contrastive} group or a
\textsc{Failed}$\rightarrow$\textsc{Solved} transition. If feedback fails to
move the group out of the \textsc{Failed} state before the turn budget is
exhausted, the prompt receives zero loss.

The GRPO branch uses a PPO-style clipped surrogate and reference-policy
regularization. The FCO branch uses a reference-normalized score, which anchors
pairwise comparisons to the fixed reference policy. This anchoring should not be
interpreted as a formal KL trust region by itself; rather, it is a practical
regularization mechanism inside the pairwise contrastive loss.

\subsection{Pseudocode}
\label{app:alg:pseudocode}

Algorithm~\ref{alg:MulFeRL} summarizes the training procedure for one prompt.
The behavior policy $\pi_{\mathrm{old}}$ is assumed to be frozen before rollout
collection for the current outer update.

\begin{algorithm*}[t]
\caption{MulFeRL training for one prompt $x$}
\label{alg:MulFeRL}
\begin{algorithmic}[1]
\REQUIRE Trainable policy $\pi_\theta$; frozen behavior policy $\pi_{\mathrm{old}}$; reference policy $\pi_{\mathrm{ref}}$; verifier $\mathcal V$; feedback provider $F_\psi$; group size $K$; max turns $T$; subgroup count $g$; GRPO parameters $(\epsilon_{\mathrm{clip}},\epsilon_{\mathrm{adv}},\beta_{\mathrm{KL}})$; FCO parameters $(\tau,\gamma)$; FCO weight $\lambda_{\mathrm{FCO}}$.
\STATE $c^{(0)}\leftarrow x$
\FOR{$t=0,\ldots,T-1$}
    \STATE Sample $\mathcal G^{(t)}=[y_i^{(t)}]_{i=1}^{K}$ with $y_i^{(t)}\sim\pi_{\mathrm{old}}(\cdot\mid c^{(t)})$
    \STATE Compute $r_i^{(t)}\leftarrow R_{\mathcal V}(x,y_i^{(t)})$ for $i=1,\ldots,K$
    \STATE Determine $z(\mathcal G^{(t)})$ by Eq.~\eqref{eq:app_reward_state}
    \IF{$z(\mathcal G^{(t)})=\textsc{Contrastive}$}
        \STATE $\mathcal L(x)\leftarrow \mathcal L_{\mathrm{GRPO}}^{(t)}$ by Eq.~\eqref{eq:app_grpo_loss}
        \STATE \textbf{return} $\mathcal L(x)$
    \ELSIF{$z(\mathcal G^{(t)})=\textsc{Solved}$}
        \IF{$t=0$}
            \STATE \textbf{return} $0$ \COMMENT{Solved without feedback; no contrastive signal}
        \ELSE
            \STATE Construct $\widehat{\mathcal Q}_{\mathrm{FCO}}^{(t)}$ from $\mathcal G^{(t)}$ and $\mathcal G^{(t-1)}$ by Eq.~\eqref{eq:app_fco_sampled_pairs}
            \STATE $\mathcal L(x)\leftarrow \lambda_{\mathrm{FCO}}\mathcal L_{\mathrm{FCO}}^{(t)}$ by Eq.~\eqref{eq:app_fco_loss}
            \STATE \textbf{return} $\mathcal L(x)$
        \ENDIF
    \ELSE
        \STATE \COMMENT{$z(\mathcal G^{(t)})=\textsc{Failed}$}
        \IF{$t=T-1$}
            \STATE \textbf{return} $0$ \COMMENT{No progress within turn budget}
        \ENDIF
        \STATE Obtain subgroup feedbacks and summarize them into $f^{(t)}$ using Section~\ref{app:alg:feedback_agg}
        \STATE $c^{(t+1)}\leftarrow \operatorname{Inject}(x,f^{(t)})$
    \ENDIF
\ENDFOR
\STATE \textbf{return} $0$
\end{algorithmic}
\end{algorithm*}

\section{Theoretical Analysis}
\label{app:theory}

We provide a theoretical analysis of MulFeRL. The goal is not to prove a global
monotonic improvement guarantee, which would require stronger assumptions on
optimization, exploration, verifier correctness, and model parameterization.
Instead, we establish a more modest but useful claim: the routing rule in
MulFeRL converts verifier-observed reward patterns into the appropriate local
training signal. In particular, GRPO extracts \emph{within-context} contrast
when successes and failures coexist in the same rollout group, whereas FCO
extracts \emph{cross-context} contrast when feedback turns an all-failed group
into an all-solved group.

This view is consistent with prior policy-optimization and preference-learning
methods. PPO optimizes a clipped surrogate objective to stabilize policy
updates~\citep{schulman2017proximal}. GRPO, introduced in DeepSeekMath, replaces
a learned value critic with group-relative normalization and has been used
successfully in verifiable-reward reasoning settings~\citep{shao2024deepseekmath,
deepseekai2025deepseekr1}. FCO is not a GRPO estimator. Rather, it is a verifier-gated
pairwise contrastive surrogate, analogous in form to reference-normalized
preference objectives such as DPO, which increase the relative score of preferred
responses over dispreferred responses under a reference-policy normalization
~\citep{rafailov2023direct}. The difference is that our preference labels are
not human preference annotations; they are induced by an automatic verifier and
a temporally adjacent Failed$\rightarrow$Solved transition.

\subsection{Preliminaries}
\label{app:theory:prelim}

For a prompt $x$ and context $c$, define the verifier-success and verifier-failure sets:

\[
S_x=\{y:R_{\mathcal V}(x,y)=1\},
\qquad
F_x=\{y:R_{\mathcal V}(x,y)=0\},
\]
The success probability under context $c$ is:

\[
p_\theta(c)
=
\Pr_{y\sim\pi_\theta(\cdot\mid c)}[y\in S_x]
\]
Here $c$ may be the original prompt or a feedback-augmented prompt. In the actual PPO/GRPO-style data-collection protocol, this rollout distribution is instantiated by the frozen behavior policy at the current outer update, i.e., $\theta=\theta_{\mathrm{old}}$; we keep the $\theta$ notation in the analysis to study the local update direction around that point.

\begin{assumption}[Conditional rollout independence]
\label{assump:independence}
Given a fixed context $c$, the $K$ rollouts in a group are sampled independently
from $\pi_\theta(\cdot\mid c)$.
\end{assumption}

Under Assumption~\ref{assump:independence}, the probabilities of the three
verifier-observed group states are:

\begin{align}
P_{\textsc{Solved}}(c)
&=
p_\theta(c)^K,
\label{eq:app_p_solved}
\\
P_{\textsc{Failed}}(c)
&=
(1-p_\theta(c))^K,
\label{eq:app_p_failed}
\\
P_{\textsc{Contrastive}}(c)
&=
1-p_\theta(c)^K-(1-p_\theta(c))^K.
\label{eq:app_p_contrastive}
\end{align}

For hard prompts, $p_\theta(c)$ is small. More precisely, if
$p_\theta(c)\ll 1$ and $Kp_\theta(c)\ll 1$, then:

\[
(1-p_\theta(c))^K
=
1-Kp_\theta(c)+O(K^2p_\theta(c)^2).
\]
The failed-state probability expands as:

\[
P_{\textsc{Failed}}(c)
=
1-Kp_\theta(c)+O(K^2p_\theta(c)^2),
\]
The contrastive-state probability expands as:

\[
P_{\textsc{Contrastive}}(c)
=
Kp_\theta(c)
-
O(K^2p_\theta(c)^2)
-
p_\theta(c)^K.
\]
This explains why hard prompts are likely to produce all-failed groups: the same
prompts for which learning is most needed are often the prompts for which
vanilla group-relative RL receives no within-group contrast.

\subsection{Uniform-Reward Groups Provide No Reward-Dependent GRPO Signal}
\label{app:theory:uniform}

\begin{proposition}[Vanishing group-relative advantage]
\label{prop:uniform_zero}
If all rewards in a rollout group are identical, then the normalized
group-relative advantages are zero. Consequently, the reward-dependent component
of the GRPO policy surrogate provides no discriminative learning signal.
\end{proposition}

\paragraph{Proof.}
Let $\mathcal G=\{y_i\}_{i=1}^{K}$ be a rollout group with rewards
$r_i=R_{\mathcal V}(x,y_i)$. GRPO-style group-relative normalization computes:

\[
\bar r=\frac{1}{K}\sum_{i=1}^{K}r_i,
\qquad
\hat\sigma_r=
\sqrt{
\frac{1}{K}\sum_{i=1}^{K}(r_i-\bar r)^2
},
\]
and the corresponding normalized advantage is:

\[
\hat A_i=\frac{r_i-\bar r}{\hat\sigma_r+\epsilon},
\]
where $\epsilon>0$ is a numerical stabilizer. If all rewards are identical, then
$r_i=\bar r$ for all $i$, and hence:

\[
\hat A_i=0,
\qquad
\forall i\in[K].
\]
The reward-dependent GRPO/PPO surrogate contains terms of the form:

\[
\min\bigl(\rho_i(\theta)\hat A_i,
\operatorname{clip}(\rho_i(\theta),1-\varepsilon,1+\varepsilon)\hat A_i\bigr),
\]
which vanish whenever $\hat A_i=0$. Therefore, uniform-reward groups provide no
reward-dependent discriminative signal. This applies to both all-failed groups
and all-solved groups.
\hfill $\square$

\paragraph{Remark.}
Proposition~\ref{prop:uniform_zero} concerns the reward-dependent part of the
policy objective. If an implementation also applies an explicit KL penalty or
other regularizer on these samples, that regularizer may still produce a
gradient. The proposition only states that no success--failure credit assignment
can be extracted from a group whose verifier rewards are uniform.

\paragraph{Implication.}
This proposition motivates the routing rule in MulFeRL. An initial
\textsc{Solved} group is skipped because the prompt is already solved and no
contrast is available. A \textsc{Failed} group triggers progress induction
because it is unsolved and low-signal. A feedback-induced \textsc{Solved} group
is not skipped, because the preceding all-failed group provides a cross-context
failure anchor.

\subsection{Progress Induction Exposes Additional Update Signals}
\label{app:theory:progress_induction}

Suppose a prompt reaches a \textsc{Failed} state at turn $t$. Feedback produces
a new context:

\[
c^{(t+1)}=\operatorname{Inject}(x,f^{(t)}).
\]
Let the pre- and post-feedback success probabilities be:

\[
p_t=p_\theta(c^{(t)}),
\qquad
q_t=p_\theta(c^{(t+1)})
\]
denote the success probabilities before and after injecting feedback.

\begin{lemma}[Escape probability after feedback]
\label{lem:escape}
Conditioned on the feedback-augmented context $c^{(t+1)}$, the probability that
the regenerated group contains at least one verifier-successful rollout is:

\[
P_{\mathrm{escape}}(c^{(t+1)})
=
1-(1-q_t)^K.
\]
Moreover, if the feedback improves the single-rollout success probability,
namely:

\[
q_t>p_t,
\]
then the escape probabilities satisfy:

\[
1-(1-q_t)^K
>
1-(1-p_t)^K.
\]
\end{lemma}

\paragraph{Proof.}
Under Assumption~\ref{assump:independence}, the regenerated group remains
all-failed with probability $(1-q_t)^K$. Therefore, the probability of escaping
the all-failed state is:

\[
P_{\mathrm{escape}}(c^{(t+1)})
=
1-(1-q_t)^K.
\]
The function $h(z)=1-(1-z)^K$ is strictly increasing for $z\in[0,1)$ and
$K\ge 1$, because:

\[
h'(z)=K(1-z)^{K-1}\ge 0,
\]
with strict positivity for $z<1$. Therefore, $q_t>p_t$ implies
$h(q_t)>h(p_t)$.
\hfill $\square$

Lemma~\ref{lem:escape} is intentionally conditional. It does not claim that any
feedback string necessarily improves the policy. It states that whenever
feedback increases the success probability from $p_t$ to $q_t$, the probability
of observing a non-failed group increases accordingly. This distinction is
important because poor feedback may leave $q_t\le p_t$.

After feedback, the regenerated group can be either \textsc{Contrastive},
\textsc{Solved}, or \textsc{Failed}. Their probabilities are:

\begin{align}
P_{\textsc{Contrastive}}(c^{(t+1)})
&=
1-q_t^K-(1-q_t)^K,
\label{eq:app_feedback_contrastive}
\\
P_{\textsc{Solved}}(c^{(t+1)})
&=
q_t^K,
\label{eq:app_feedback_solved}
\\
P_{\textsc{Failed}}(c^{(t+1)})
&=
(1-q_t)^K.
\label{eq:app_feedback_failed}
\end{align}

Vanilla GRPO can use the \textsc{Contrastive} case because successes and failures
coexist in the same group. However, by Proposition~\ref{prop:uniform_zero}, it
cannot extract reward-dependent success--failure credit from the
\textsc{Solved} case because all rewards are equal. MulFeRL additionally uses
the feedback-induced \textsc{Solved} case through FCO. Hence, conditioned on
entering a feedback turn, the probability that MulFeRL obtains a verifier-based
update signal is:

\begin{align}
P_{\mathrm{update}}^{\mathrm{MulFeRL}}
&=
P_{\textsc{Contrastive}}(c^{(t+1)})
\\
&\quad+
P_{\textsc{Solved}}(c^{(t+1)})
\nonumber\\
&=
1-(1-q_t)^K.
\label{eq:app_update_mulferl}
\end{align}
The corresponding probability for GRPO after the same feedback turn is:

\[
\begin{aligned}
P_{\mathrm{update}}^{\mathrm{GRPO}}
&=
P_{\textsc{Contrastive}}(c^{(t+1)})
\\
&=
1-q_t^K-(1-q_t)^K.
\end{aligned}
\]
Therefore, the update probability recovered by FCO is:

\begin{equation}
\begin{aligned}
P_{\mathrm{update}}^{\mathrm{MulFeRL}}
&-
P_{\mathrm{update}}^{\mathrm{GRPO}}
\\
&=
q_t^K.
\end{aligned}
\label{eq:app_fco_recovered_mass}
\end{equation}

Equation~\eqref{eq:app_fco_recovered_mass} characterizes the precise probability
mass that is lost by a purely within-group contrastive method but recovered by
FCO: the feedback-induced all-solved groups.

\subsection{GRPO Provides Within-Context Success--Failure Credit Assignment}
\label{app:theory:grpo_direction}

We next analyze the local reward-gradient direction of binary-reward GRPO on a
\textsc{Contrastive} group. Consider a group:

\[
\mathcal G=\mathcal G^+\cup\mathcal G^-,
\]
where the group sizes are:

\[
\begin{aligned}
\mathcal G^+
&=\{y_i:R_{\mathcal V}(x,y_i)=1\},\\
\mathcal G^-
&=\{y_i:R_{\mathcal V}(x,y_i)=0\}.
\end{aligned}
\]
Define:

\[
m=|\mathcal G^+|,
\qquad
n=|\mathcal G^-|,
\qquad
K=m+n,
\]
with $m>0$ and $n>0$.

For binary rewards:

\[
\bar r=\frac{m}{K},
\qquad
\hat\sigma_r=
\sqrt{\frac{m}{K}\frac{n}{K}}
=
\frac{\sqrt{mn}}{K}.
\]
With numerical stabilizer $\epsilon>0$, define:

\[
d_\epsilon=\frac{\sqrt{mn}}{K}+\epsilon.
\]
Then the normalized advantage for a successful rollout is:

\[
\hat A^+
=
\frac{1-\bar r}{d_\epsilon}
=
\frac{n/K}{d_\epsilon},
\]
and the normalized advantage for a failed rollout is:

\[
\hat A^-
=
\frac{0-\bar r}{d_\epsilon}
=
-\frac{m/K}{d_\epsilon}.
\]

\begin{proposition}[Local GRPO contrast direction]
\label{prop:grpo_direction}
At $\theta=\theta_{\mathrm{old}}$, ignoring clipping saturation and explicit
regularization terms, the reward-dependent GRPO gradient on a binary-reward
\textsc{Contrastive} group is a positive scalar multiple of:

\[
\begin{aligned}
&\frac{1}{m}\sum_{y^+\in\mathcal G^+}
\nabla_\theta\log\pi_\theta(y^+\mid c)
\\
&\quad-
\frac{1}{n}\sum_{y^-\in\mathcal G^-}
\nabla_\theta\log\pi_\theta(y^-\mid c).
\end{aligned}
\]
\end{proposition}

\paragraph{Proof.}
The unclipped reward-dependent local surrogate has gradient:

\[
\begin{aligned}
\nabla_\theta J_{\mathrm{GRPO}}
&=
\frac{1}{K}
\Biggl[
\sum_{y^+\in\mathcal G^+}
\hat A^+
\nabla_\theta\log\pi_\theta(y^+\mid c)
\\
&\quad+
\sum_{y^-\in\mathcal G^-}
\hat A^-
\nabla_\theta\log\pi_\theta(y^-\mid c)
\Biggr],
\end{aligned}
\]
because $\nabla_\theta \rho_i(\theta)|_{\theta=\theta_{\mathrm{old}}}
=\nabla_\theta\log\pi_\theta(y_i\mid c)$.
Substituting the binary advantages gives:

\[
\begin{aligned}
\nabla_\theta J_{\mathrm{GRPO}}
&=
\frac{1}{K}
\Biggl[
\frac{n/K}{d_\epsilon}
\sum_{y^+\in\mathcal G^+}
\nabla_\theta\log\pi_\theta(y^+\mid c)
\\
&\quad-
\frac{m/K}{d_\epsilon}
\sum_{y^-\in\mathcal G^-}
\nabla_\theta\log\pi_\theta(y^-\mid c)
\Biggr].
\end{aligned}
\]
Equivalently:

\[
\begin{aligned}
\nabla_\theta J_{\mathrm{GRPO}}
&=
\frac{mn}{K^2d_\epsilon}
\Biggl[
\frac{1}{m}
\sum_{y^+\in\mathcal G^+}
\nabla_\theta\log\pi_\theta(y^+\mid c)
\\
&\quad-
\frac{1}{n}
\sum_{y^-\in\mathcal G^-}
\nabla_\theta\log\pi_\theta(y^-\mid c)
\Biggr].
\end{aligned}
\]
Since $m,n,K,d_\epsilon>0$, the scalar multiplier is positive.
\hfill $\square$

Proposition~\ref{prop:grpo_direction} shows that, locally and on the
reward-dependent part of the objective, GRPO increases the average log-likelihood
of verifier-successful samples relative to verifier-failed samples under the
same context. This is a within-context credit-assignment signal. The statement
is local: clipping, KL regularization, parameter sharing, finite step sizes, and
optimization noise can affect the realized update.

\subsection{FCO Provides Cross-Context Success--Failure Credit Assignment}
\label{app:theory:fco_direction}

We now analyze the case that motivates FCO. Consider a
\textsc{Failed}$\rightarrow$\textsc{Solved} transition. Let
$\mathcal G^{(t-1)}$ be the all-failed group sampled under context
$c^{(t-1)}$, and let $\mathcal G^{(t)}$ be the all-solved group sampled under
the feedback-augmented context $c^{(t)}$. Thus, for all
$y^-\in\mathcal G^{(t-1)}$ and $y^+\in\mathcal G^{(t)}$:

\[
R_{\mathcal V}(x,y^-)=0,
\qquad
R_{\mathcal V}(x,y^+)=1.
\]

FCO defines the length-normalized reference score:

\[
s_\theta(c,y)
=
\frac{1}{|y|}
\left[
\log\pi_\theta(y\mid c)
-
\log\pi_{\mathrm{ref}}(y\mid c)
\right],
\]
where $\pi_{\mathrm{ref}}$ is fixed. For a positive--negative pair
$(y^+,y^-)$, define the cross-context score gap:

\[
D_\theta(y^+,y^-)
=
s_\theta(c^{(t)},y^+)
-
s_\theta(c^{(t-1)},y^-).
\]
The pairwise FCO loss is:

\[
\ell_{\mathrm{FCO}}(y^+,y^-)
=
-\log\sigma\left(
\tau\left[D_\theta(y^+,y^-)-\gamma\right]
\right),
\]
where $\tau>0$ is an inverse-temperature parameter and $\gamma\ge 0$ is a
margin.

\begin{proposition}[Local FCO contrast direction]
\label{prop:fco_direction}
For any verifier-certified pair $(y^+,y^-)$ from a
\textsc{Failed}$\rightarrow$\textsc{Solved} transition, the negative gradient
direction of the FCO loss is:

\[
\begin{aligned}
-\nabla_\theta \ell_{\mathrm{FCO}}(y^+,y^-)
&=
\alpha_\theta(y^+,y^-)
\Biggl[
\frac{1}{|y^+|}
\nabla_\theta\log\pi_\theta(y^+\mid c^{(t)})
\\
&\quad-
\frac{1}{|y^-|}
\nabla_\theta\log\pi_\theta(y^-\mid c^{(t-1)})
\Biggr],
\end{aligned}
\]
where the scalar coefficient is:

\[
\alpha_\theta(y^+,y^-)
=
\tau\sigma\left(
-\tau\left[D_\theta(y^+,y^-)-\gamma\right]
\right)
\]
satisfies:

\[
0<\alpha_\theta(y^+,y^-)<\tau.
\]
\end{proposition}

\paragraph{Proof.}
Let the scalar logit be:

\[
z_\theta
=
\tau\left[D_\theta(y^+,y^-)-\gamma\right].
\]
Since:

\[
\ell_{\mathrm{FCO}}=-\log\sigma(z_\theta),
\]
we have:

\[
\begin{aligned}
-\nabla_\theta \ell_{\mathrm{FCO}}
&=
\sigma(-z_\theta)\nabla_\theta z_\theta
\\
&=
\tau\sigma(-z_\theta)\nabla_\theta D_\theta(y^+,y^-).
\end{aligned}
\]
Because $\pi_{\mathrm{ref}}$ is fixed:

\[
\nabla_\theta s_\theta(c,y)
=
\frac{1}{|y|}
\nabla_\theta\log\pi_\theta(y\mid c).
\]
Therefore, the score-gradient is:

\[
\begin{aligned}
\nabla_\theta D_\theta(y^+,y^-)
&=
\frac{1}{|y^+|}
\nabla_\theta\log\pi_\theta(y^+\mid c^{(t)})
\\
&\quad-
\frac{1}{|y^-|}
\nabla_\theta\log\pi_\theta(y^-\mid c^{(t-1)}).
\end{aligned}
\]
Finally, since $\sigma(\cdot)\in(0,1)$ and $\tau>0$, the scalar coefficient
$\alpha_\theta(y^+,y^-)$ lies in $(0,\tau)$.
\hfill $\square$

Proposition~\ref{prop:fco_direction} shows that FCO reinforces
verifier-successful responses under the feedback-augmented context while
suppressing verifier-failed responses under the preceding failed context. This
is a cross-context contrastive signal. Unlike GRPO, it is not an unbiased
estimator of a same-context policy gradient, because the positive and negative
samples are conditioned on different contexts. Its role is instead to convert a
verifier-certified temporal transition into a pairwise training signal.

\subsection{Unified View: Where the Contrast Is Observable}
\label{app:theory:unified}

The preceding propositions show that GRPO and FCO have the same local sign
structure but operate on different observable contrasts:

\[
\begin{gathered}
\text{increase verifier-successful samples}\\
\text{relative to verifier-failed samples}.
\end{gathered}
\]
The distinction is where this success--failure contrast is observed.

In a \textsc{Contrastive} group, successes and failures coexist under the same
context. GRPO can therefore use within-context group-relative advantages. In a
\textsc{Failed}$\rightarrow$\textsc{Solved} transition, the current group is
uniformly positive and has no within-group reward variation. GRPO is
uninformative by Proposition~\ref{prop:uniform_zero}. However, the immediately
preceding all-failed group provides a verifier-certified negative anchor, and
FCO uses the adjacent cross-context contrast.

Thus, MulFeRL does not combine unrelated losses. It routes each prompt according
to the location of the verifier-observed discrimination signal:

\[
\begin{cases}
\textsc{Contrastive}: & \text{GRPO within-context},\\
\textsc{Failed}\rightarrow\textsc{Solved}: &
\text{FCO cross-context},\\
\textsc{Failed}\rightarrow\textsc{Failed}: &
\text{no positive update},\\
\textsc{Solved at initial turn}: &
\text{skip}.
\end{cases}
\]

This routing is also consistent with the broader feedback-refinement literature:
language feedback can change the conditioning context and improve subsequent
attempts without itself being treated as a reward label~\citep{madaan2023self,
shinn2023reflexion}. MulFeRL uses feedback only to induce a new context; the
training signal is still gated by the verifier.

\subsection{Why FCO Is Needed for Reward-Saturated Progress}
\label{app:theory:why_fco}

A feedback-induced \textsc{Solved} group is a reward-saturated progress event.
It is useful behaviorally because all regenerated candidates are correct.
However, it is unusable for reward-dependent GRPO because all normalized
advantages vanish. If such groups were skipped, the algorithm would discard
precisely the cases in which feedback was most effective.

FCO recovers this otherwise discarded signal. Since feedback is triggered only
after a \textsc{Failed} group, every feedback-induced \textsc{Solved} group has
an adjacent all-failed anchor:

\[
\mathcal G^{(t-1)}:\text{all failed},
\qquad
\mathcal G^{(t)}:\text{all solved}.
\]
The verifier therefore certifies every cross pair:

\[
(y^+,y^-)\in\mathcal G^{(t)}\times\mathcal G^{(t-1)}
\]
as ordered in the desired direction:

\[
R_{\mathcal V}(x,y^+)=1
>
0=R_{\mathcal V}(x,y^-).
\]
FCO turns this ordered transition into a pairwise contrastive objective. It is
therefore not an auxiliary preference loss over arbitrary model generations; it
is the mechanism that makes reward-saturated progress learnable.

\subsection{Unbiasedness of Permutation-Based FCO Pair Sampling}
\label{app:theory:pair_sampling}

The full FCO objective over a transition averages over all $K^2$ positive--
negative pairs:

\[
\mathcal L_{\mathrm{full}}
=
\frac{1}{K^2}
\sum_{j=1}^{K}
\sum_{i=1}^{K}
\ell_{\mathrm{FCO}}(y_j^+,y_i^-).
\]
To reduce computation, MulFeRL samples a random permutation $\pi$ over
$\{1,\ldots,K\}$ and uses:

\[
\mathcal L_{\mathrm{perm}}
=
\frac{1}{K}
\sum_{j=1}^{K}
\ell_{\mathrm{FCO}}(y_j^+,y_{\pi(j)}^-).
\]

\begin{proposition}[Unbiased pair sampling]
\label{prop:perm_unbiased}
If $\pi$ is sampled uniformly at random from all permutations of
$\{1,\ldots,K\}$, then:

\[
\mathbb E_\pi[\mathcal L_{\mathrm{perm}}]
=
\mathcal L_{\mathrm{full}}.
\]
Moreover, whenever differentiation and expectation can be interchanged:

\[
\mathbb E_\pi[\nabla_\theta\mathcal L_{\mathrm{perm}}]
=
\nabla_\theta\mathcal L_{\mathrm{full}}.
\]
\end{proposition}

\paragraph{Proof.}
For any fixed $j$, $\pi(j)$ is uniformly distributed over
$\{1,\ldots,K\}$. Therefore:

\[
\mathbb E_\pi
\left[
\ell_{\mathrm{FCO}}(y_j^+,y_{\pi(j)}^-)
\right]
=
\frac{1}{K}
\sum_{i=1}^{K}
\ell_{\mathrm{FCO}}(y_j^+,y_i^-).
\]
Averaging over $j$ gives:

\[
\mathbb E_\pi[\mathcal L_{\mathrm{perm}}]
=
\frac{1}{K}
\sum_{j=1}^{K}
\frac{1}{K}
\sum_{i=1}^{K}
\ell_{\mathrm{FCO}}(y_j^+,y_i^-)
=
\mathcal L_{\mathrm{full}}.
\]
The gradient statement follows by linearity of expectation under the standard
regularity condition that the gradient can be exchanged with expectation.
\hfill $\square$

This result shows that permutation-based FCO preserves the full all-pair
objective in expectation while reducing the number of evaluated pairs from
$K^2$ to $K$.

\subsection{Verifier Gating and Regularization}
\label{app:theory:stability}

MulFeRL is designed to avoid directly optimizing noisy feedback text. Feedback
only changes the next-turn conditioning context. A policy update is applied only
when the verifier observes either a \textsc{Contrastive} group or a
\textsc{Failed}$\rightarrow$\textsc{Solved} transition. Therefore, unsuccessful
feedback does not create positive labels by itself.

The update is also regularized in two ways. First, the GRPO branch follows
PPO-style clipped policy optimization and, in common implementations, includes a
reference-policy KL term~\citep{schulman2017proximal,shao2024deepseekmath}.
These mechanisms are intended to limit large policy deviations during on-policy
optimization. Second, FCO uses the reference-normalized score:

\[
s_\theta(c,y)
=
\frac{1}{|y|}
\left[
\log\pi_\theta(y\mid c)
-
\log\pi_{\mathrm{ref}}(y\mid c)
\right].
\]
This score anchors the pairwise comparison to changes relative to the reference
policy: if $\pi_\theta=\pi_{\mathrm{ref}}$, then $s_\theta(c,y)=0$ for all
$(c,y)$. Hence FCO prefers a positive response over a negative response only to
the extent that the current policy improves their relative normalized score
beyond the reference baseline.

This reference normalization should not be interpreted as a formal KL trust
region by itself. It is an anchoring mechanism inside a pairwise logistic loss.
A formal global stability guarantee would require additional assumptions, such
as explicit KL constraints, bounded gradients, sufficiently small step sizes, or
convexity-like conditions that do not hold for neural language models. Our claim
is therefore limited to verifier-gated, reference-anchored local optimization.

\subsection{Summary of the Theoretical Justification}
\label{app:theory:summary}

The analysis establishes the following claims.

First, uniform-reward groups provide no reward-dependent GRPO credit assignment:
their group-relative advantages vanish. This explains why all-failed and
all-solved groups are uninformative for purely within-group contrastive RL.

Second, feedback-guided regeneration exposes additional update opportunities.
Conditioned on a feedback-induced success probability $q_t$, the probability of
escaping an all-failed state is $1-(1-q_t)^K$. If feedback improves the
single-rollout success probability from $p_t$ to $q_t>p_t$, this escape
probability strictly increases. Among the escaped cases, FCO recovers exactly
the all-solved probability mass $q_t^K$ that GRPO cannot use.

Third, GRPO and FCO share the same local sign structure: both increase
verifier-successful samples relative to verifier-failed samples. The difference
is where the contrast is observed. GRPO uses within-context contrast in a
\textsc{Contrastive} group. FCO uses adjacent cross-context contrast in a
\textsc{Failed}$\rightarrow$\textsc{Solved} transition.

Finally, FCO should be viewed as a verifier-certified cross-context contrastive
surrogate, not as an unbiased estimator of the original GRPO objective. This
interpretation is both mathematically accurate and sufficient to justify the
algorithmic role of FCO: it turns reward-saturated progress events into usable
training signals while preserving verifier gating and reference anchoring.

\section{Prompts for MulFeRL}
\label{app:prompts}
\begin{figure*}[t]
\centering
\begin{tcolorbox}[
  title=Training system prompt,
  colback=boxbg,
  colframe=black,
  coltitle=white,
  colbacktitle=titlebg,
  fonttitle=\bfseries,
  sharp corners=south,
  boxrule=0.8pt,
  width=\textwidth
]
\small
{\ttfamily
You are a reasoning assistant.\textbackslash n\\
Solve the problem step by step.\textbackslash n\textbackslash n\\
You MUST follow this exact output format for every problem:\textbackslash n\\
1) Wrap ALL your reasoning inside a single <thinking>...</thinking> block.\textbackslash n\\
2) At the very start of <thinking>, output exactly ONE <feedback>...</feedback> block.\textbackslash n\\
\ \ \ - The <feedback> must be concise and actionable, and use this structure:\textbackslash n\\
\ \ \ \ \ <feedback>\textbackslash n\\
\ \ \ \ \ Issue:\textbackslash n\\
\ \ \ \ \ 1. Likely pitfalls: ...\textbackslash n\\
\ \ \ \ \ 2. Step-by-step plan:\textbackslash n\\
\ \ \ \ \ \ \ \ - (1) ...\textbackslash n\\
\ \ \ \ \ \ \ \ - (2) ...\textbackslash n\\
\ \ \ \ \ \ \ \ - (3) ...\textbackslash n\textbackslash n\\
\ \ \ \ \ Fix steps:\textbackslash n\\
\ \ \ \ \ 1. ...\textbackslash n\\
\ \ \ \ \ 2. ...\textbackslash n\\
\ \ \ \ \ 3. ...\textbackslash n\\
\ \ \ \ \ </feedback>\textbackslash n\\
\ \ \ - Give guidance / a repair plan, but do NOT give a full solution inside <feedback>.\textbackslash n\\
\ \ \ - Do NOT output any expression inside <feedback> that directly equals the final result.\textbackslash n\\
\ \ \ \ \ (e.g., do NOT write something like '... = <final answer>' inside <feedback>.)\textbackslash n\\
\ \ \ - You MAY include tiny snippets (a short identity, a one-line correction),\textbackslash n\\
\ \ \ \ \ but avoid long derivations or long equations in <feedback>.\textbackslash n\\
3) After </thinking>, on a new line, output the final numeric answer in the format:\textbackslash n\\
\ \ \ \textbackslash boxed\{answer\}\textbackslash n\\
Do NOT add any extra text after the boxed answer.\textbackslash n\textbackslash n\\
Example (format only):\textbackslash n\\
<thinking>\textbackslash n\\
<feedback>\textbackslash n\\
Issue:\textbackslash n\\
1. Likely pitfalls: Misreading quantities; forgetting to combine changes.\textbackslash n\\
2. Step-by-step plan:\textbackslash n\\
\ \ \ - (1) Identify the initial quantity and each change.\textbackslash n\\
\ \ \ - (2) Choose the correct operation (add/subtract/etc.).\textbackslash n\\
\ \ \ - (3) Compute carefully.\textbackslash n\textbackslash n\\
Fix steps:\textbackslash n\\
1. Extract numbers and what they represent.\textbackslash n\\
2. Write the operation clearly.\textbackslash n\\
3. Recompute the final arithmetic once.\textbackslash n\\
</feedback>\textbackslash n\\
Alice starts with 3 apples.\textbackslash n\\
She buys 2 more apples.\textbackslash n\\
Total apples = 3 + 2 = 5.\textbackslash n\\
</thinking>\textbackslash n\\
\textbackslash boxed\{5\}\textbackslash n\\
}
\end{tcolorbox}

\caption{Training system prompt.}
\label{fig:input_pem}
\end{figure*}

\begin{figure*}[t]
\centering
\begin{tcolorbox}[
  title=Obtain Feedback system prompt,
  colback=boxbg,
  colframe=black,
  coltitle=white,
  colbacktitle=titlebg,
  fonttitle=\bfseries,
  sharp corners=south,
  boxrule=0.8pt,
  width=\textwidth
]
\small
{\ttfamily
You are a strict reviewer of an incorrect solution.\textbackslash n\\
Please briefly explain (step-by-step) the solution's mistakes and provide your correction suggestions.\textbackslash n\textbackslash n\\
Task:\textbackslash n\\
- Identify the earliest/root mistake (the first wrong step).\textbackslash n\\
- Explain the errors step-by-step (concise, specific).\textbackslash n\\
- Give concrete correction suggestions (how to fix the process).\textbackslash n\textbackslash n\\
Rules:\textbackslash n\\
- Do NOT provide a full correct solution.\textbackslash n\\
- Do NOT provide or reveal the final numeric answer.\textbackslash n\\
- Avoid generic advice; point to specific steps or lines.\textbackslash n\textbackslash n\\
Output ONLY:\textbackslash n\\
<feedback>\textbackslash n\\
Issue:\textbackslash n\\
1. Earliest/root mistake: ...\textbackslash n\\
2. Where it first goes wrong (quote 1--2 lines): "\ldots"\textbackslash n\\
3. Step-by-step errors:\textbackslash n\\
\ \ \ - (1) ...\textbackslash n\\
\ \ \ - (2) ...\textbackslash n\\
\ \ \ - (3) ...\textbackslash n\textbackslash n\\
Fix steps:\textbackslash n\\
1. ...\textbackslash n\\
2. ...\textbackslash n\\
3. ...\textbackslash n\\
</feedback>%
}
\end{tcolorbox}

\caption{System prompt for obtaining feedback.}
\label{fig:feedback_system_prompt}
\end{figure*}

\begin{figure*}[t]
\centering
\begin{tcolorbox}[
  title=Merge feedback system prompt,
  colback=boxbg,
  colframe=black,
  coltitle=white,
  colbacktitle=titlebg,
  fonttitle=\bfseries,
  sharp corners=south,
  boxrule=0.8pt,
  width=\textwidth
]
\small
{\ttfamily
You merge multiple feedback comments on an incorrect  solution.\textbackslash n\textbackslash n\\
Task:\textbackslash n\\
- Combine the feedback into ONE concise, actionable feedback.\textbackslash n\\
- Deduplicate repeated points.\textbackslash n\\
- Keep ONLY the earliest/root failure modes (the first wrong step and its consequences).\textbackslash n\textbackslash n\\
Rules:\textbackslash n\\
- Do NOT provide a full correct solution.\textbackslash n\\
- Do NOT provide or reveal the final numeric answer.\textbackslash n\\
- Keep it short and specific.\textbackslash n\textbackslash n\\
Output ONLY:\textbackslash n\\
<feedback>\textbackslash n\\
Issue:\textbackslash n\\
1. Earliest/root mistake: ...\textbackslash n\\
2. Where it first goes wrong (quote 1--2 lines): "\ldots"\textbackslash n\\
3. Step-by-step errors:\textbackslash n\\
\ \ \ - (1) ...\textbackslash n\\
\ \ \ - (2) ...\textbackslash n\\
\ \ \ - (3) ...\textbackslash n\textbackslash n\\
Fix steps:\textbackslash n\\
1. ...\textbackslash n\\
2. ...\textbackslash n\\
3. ...\textbackslash n\\
</feedback>%
}
\end{tcolorbox}

\caption{System prompt for merging feedback.}
\label{fig:merge_feedback_system_prompt}
\end{figure*}

\begin{figure*}[t]
\centering
\begin{tcolorbox}[
  title=Feedback injection and regeneration rollout prompt,
  colback=boxbg,
  colframe=black,
  coltitle=white,
  colbacktitle=titlebg,
  fonttitle=\bfseries,
  sharp corners=south,
  boxrule=0.8pt,
  width=\textwidth
]
\small
{\ttfamily
You are a problem reasoning assistant.\textbackslash n\\
Re-solve the problem from scratch. Use the provided feedback only as hints.\textbackslash n\\
Do NOT create another <feedback> block.\textbackslash n\\
Do NOT modify, repeat, or paraphrase the provided feedback text.\textbackslash n\textbackslash n\\
Output format (must follow exactly):\textbackslash n\\
1) Start with EXACTLY: <thinking><feedback>\{feedback\}</feedback>\textbackslash n\\
2) Continue your reasoning, then close: </thinking>\textbackslash n\\
3) The final answer MUST be written as: \textbackslash boxed\{\{answer\}\}\textbackslash n\\
Do NOT output anything else.\textbackslash n\textbackslash n\\
Problem:\textbackslash n\\
\{question\}\textbackslash n\\
Start your new solution by continuing from the following prefix exactly:\textbackslash n\\
<thinking><feedback>\{feedback\}</feedback>%
}
\end{tcolorbox}

\caption{Prompt for feedback injection and regeneration rollout.}
\label{fig:feedback_injection_regen_prompt}
\end{figure*}

We use a small set of prompts throughout training. Concretely, we include three system prompts (for training rollouts, feedback generation, and feedback merging) and one prompt template for feedback injection and regeneration. We provide the exact prompt texts in Figures~\ref{fig:input_pem}, \ref{fig:feedback_system_prompt}, \ref{fig:merge_feedback_system_prompt}, and \ref{fig:feedback_injection_regen_prompt}, respectively. The corresponding user prompts are simple and self-explanatory (e.g., the problem statement, and when needed, the model’s sampled rollouts and/or multiple feedback comments), so we omit them to avoid redundancy.

During evaluation, for experiments that require feedback, we use the same input system prompt, feedback-generation prompt, and feedback-injection/regeneration prompt for MulFeRL on math benchmarks as those used in training, ensuring consistency with the training setup and facilitating reproducibility. Feedback merging is not required in this setting because these evaluation experiments do not aggregate feedback from multiple sources or multiple feedback comments; instead, they use a single-turn, feedback-guided regeneration protocol. For OOD evaluations, we only adjust the output-format constraints to match each task’s requirements, while keeping the overall prompt structure and interaction protocol the same. In the appendix, we present the prompts for math tasks as a representative example.

\subsection{Training System Prompt}
\label{app:prompts:system:train}
This system prompt specifies the base role and output schema for policy rollouts during training: the model must wrap all reasoning inside a single \texttt{<thinking>} block and include a single \texttt{<feedback>} block at the beginning of \texttt{<thinking>} to standardize the response structure. This schema is also used by the verifier for format compliance. See Figure~\ref{fig:input_pem}.

\subsection{Feedback Simulator System Prompt}
\label{app:prompts:system:feedback_provider}
This system prompt is used by the feedback simulator $\mathcal{F}_\psi$ and is only triggered in the \textsc{FAILED} regime (Algorithm~\ref{alg:MulFeRL}, lines 6--15). Given the original prompt $x$ and a set of failed rollouts (we pass the \emph{full} model outputs, including tags such as \texttt{<thinking>}, \texttt{<feedback>}, and the final answer field), the feedback simulator produces one concise \texttt{<feedback>} block that (i) pinpoints the earliest/root mistake, (ii) quotes where the reasoning first goes wrong, and (iii) provides step-by-step correction suggestions. Importantly, it must not provide a full corrected solution or reveal the final numeric answer. In our two-step aggregation (Appendix~\ref{app:alg:feedback_agg}), we apply this prompt to each subgroup to obtain subgroup feedbacks $\{f^{(t)}_j\}_{j=1}^g$. See Figure~\ref{fig:feedback_system_prompt}.

\subsection{Merge Feedback System Prompt}
\label{app:prompts:system:merge_feedback}
This system prompt implements the second step of feedback aggregation (Appendix~\ref{app:alg:feedback_agg}): it summarizes and merges multiple subgroup feedback comments $\{f^{(t)}_j\}_{j=1}^g$ into a single concise and actionable feedback $f^{(t)}$ with the same (Issue, Fix steps) structure used for injection. It deduplicates repeated points and keeps only the earliest/root failure modes and their immediate consequences, without providing a full solution or revealing the final answer. See Figure~\ref{fig:merge_feedback_system_prompt}.

\subsection{Feedback Injection and Regeneration Prompt}
\label{app:prompts:feedback_injection_regen}
This prompt template is used for feedback-guided regeneration when transitioning from turn $t$ to $t{+}1$ (Algorithm~\ref{alg:MulFeRL}, lines 12--14). It prepends the external feedback verbatim and forces the model to continue generation from the exact prefix \texttt{<thinking><feedback>\{feedback\}</feedback>}, ensuring the next rollout conditions on $f^{(t)}$ under standard next-token prediction. The template also forbids generating a new feedback block or paraphrasing the provided feedback, so that the injected feedback span is well-defined for token masking during optimization (Appendix~\ref{app:alg:inject_mask}). See Figure~\ref{fig:feedback_injection_regen_prompt}.

\begin{figure*}[t]
\centering
\begin{tcolorbox}[
  title=SFT/RAFT system prompt and CITL-FT initial prompt,
  colback=boxbg,
  colframe=black,
  coltitle=white,
  colbacktitle=titlebg,
  fonttitle=\bfseries,
  sharp corners=south,
  boxrule=0.8pt,
  width=\textwidth
]
\small
{\ttfamily
You are a reasoning assistant.\textbackslash n\\
Solve the problem step by step.\textbackslash n\textbackslash n\\
Output format (must follow exactly):\textbackslash n\\
1) Wrap ALL reasoning inside a single <thinking>...</thinking> block.\textbackslash n\\
2) After </thinking>, on a new line, output the final numeric answer in the format:\textbackslash n\\
\ \ \ \textbackslash boxed\{answer\}\textbackslash n\\
Do NOT output any extra text after the boxed answer.\textbackslash n
}
\end{tcolorbox}
\caption{SFT/RAFT prompt and CITL-FT initial prompt}
\label{fig:sft_ini}
\end{figure*}

\begin{figure*}[t]
\centering
\begin{tcolorbox}[
  title=CITL-FT critique system prompt,
  colback=boxbg,
  colframe=black,
  coltitle=white,
  colbacktitle=titlebg,
  fonttitle=\bfseries,
  sharp corners=south,
  boxrule=0.8pt,
  width=\textwidth
]
\small
{\ttfamily
You are a strict reviewer of a solution.\textbackslash n\\
Given a problem and a proposed solution, identify mistakes and provide actionable repair suggestions.\textbackslash n\textbackslash n\\
Rules:\textbackslash n\\
- Do NOT provide a full correct solution.\textbackslash n\\
- Do NOT reveal the final numeric answer.\textbackslash n\\
- Be specific (point to the first wrong step and its consequence).\textbackslash n\textbackslash n\\
Output ONLY:\textbackslash n\\
<critique>\textbackslash n\\
1. Earliest/root mistake: ...\textbackslash n\\
2. Why it is wrong: ...\textbackslash n\\
3. Fix plan (high-level):\textbackslash n\\
\ \ \ - (1) ...\textbackslash n\\
\ \ \ - (2) ...\textbackslash n\\
\ \ \ - (3) ...\textbackslash n\\
</critique>\textbackslash n
}
\end{tcolorbox}
\caption{Critique prompt for CITL-FT}
\label{fig:citl_critique}
\end{figure*}

\begin{figure*}[t]
\centering
\begin{tcolorbox}[
  title=CITL-FT refinement system prompt,
  colback=boxbg,
  colframe=black,
  coltitle=white,
  colbacktitle=titlebg,
  fonttitle=\bfseries,
  sharp corners=south,
  boxrule=0.8pt,
  width=\textwidth
]
\small
{\ttfamily
You are a reasoning assistant.\textbackslash n\\
Re-solve the problem, using the provided critique as guidance.\textbackslash n\\
Do NOT repeat the critique verbatim.\textbackslash n\textbackslash n\\
Output format (must follow exactly):\textbackslash n\\
1) <thinking> ... </thinking>\textbackslash n\\
2) \textbackslash boxed\{answer\}\textbackslash n\\
Do NOT output anything else.\textbackslash n\textbackslash n\\
Inputs:\textbackslash n\\
Problem: \{question\}\textbackslash n\\
Initial solution: \{initial\_response\}\textbackslash n\\
Critique: \{critique\}\textbackslash n
}
\end{tcolorbox}
\caption{Refinement prompt for CITL-FT}
\label{fig:citl_refinement}
\end{figure*}

\section{Implementation Details}
\label{app:hparams}
This section reports the key implementation choices and key hyperparameters used in our experiments.
Our implementation is built on the \texttt{verl} \citep{sheng2025hybridflow} framework.
We only report the most influential hyperparameters here; the full configuration can be found in the code repository referenced in the main text.

Our experiments use publicly available reasoning benchmarks, including math, science, and general reasoning tasks, and do not involve collecting new personal data. These benchmarks are not intended to contain personally identifying information. The generated feedback is restricted to task-level error analysis and does not require or introduce personal information.

\subsection{Training Details}
\subsubsection{RL-Training Details}
\label{app:hparams:rl}
Table~\ref{tab:rl_details} summarizes the key hyperparameters for MulFeRL. We mainly employ GPT-4o as the feedback simulator $\psi$.
For fair comparison, all reinforcement learning-based finetuning baselines share the same backbone, training data, rollout group size, sampling temperature, length limits, number of training steps, and feedback provider whenever applicable. Method-specific objectives are kept faithful to the original algorithms.

For Critique-GRPO~\citep{zhang2025critique}, we also use GPT-4o as the feedback simulator and follow the original setup, sampling 7 rollouts in the first turn and 1 rollout in the second turn.
For RLTF-FM~\citep{song2026expanding}, we follow the feedback-modeling variant of RL from Text Feedback. RLTF assumes that text feedback is available during training but not at inference, so the model must internalize feedback into its single-turn behavior. RLTF-FM implements this by adding an auxiliary objective that predicts the feedback text, i.e., the critique, conditioned on the prompt and sampled response, in addition to the standard RL objective. We instantiate this objective with GPT-4o-generated feedback and use the same rollout budget and feedback budget as other feedback-based baselines. The feedback-modeling loss is applied only to the feedback tokens, with coefficient $\lambda_{\mathrm{FM}}=0.2$. At test time, RLTF-FM is evaluated in the same single-pass setting without external feedback unless explicitly stated otherwise.

We detail MulFeRL’s prompting pipeline as follows: the initial rollout prompt is given in Figure~\ref{fig:input_pem}; the feedback-request prompt used to query $\psi$ is in Figure~\ref{fig:feedback_system_prompt}, the feedback-merge (application) prompt is in Figure~\ref{fig:merge_feedback_system_prompt}; and the prompt used for feedback injection and feedback-conditioned regeneration is in Figure~\ref{fig:feedback_injection_regen_prompt}. For other RL baselines, including GRPO, Dr.GRPO, Critique-GRPO, and RLTF-FM, we do not enforce MulFeRL’s structured output format; instead, we follow their default generation interface and, when applicable, use the prompting setup from previous work~\citep{zhang2025critique}. During training, MulFeRL receives reward $1$ only when both the output format and the final answer are correct (and $0$ otherwise), whereas for other RL methods the reward is $1$ if the final answer is correct (and $0$ otherwise).

\begin{table}[t]
\centering
\small
\setlength{\tabcolsep}{7pt}
\renewcommand{\arraystretch}{1.10}
\begin{tabularx}{\linewidth}{@{}X r@{}}
\toprule
\textbf{Parameter} & \textbf{Value} \\
\midrule

\multicolumn{2}{@{}l}{\textbf{Optimization}}\\
\addlinespace[1pt]
Global batch size & $16$ \\
Per-device micro-batch & $2$ \\
Learning rate & $1\times 10^{-6}$ \\
Total steps & $400$ \\
\midrule

\multicolumn{2}{@{}l}{\textbf{GRPO / KL / Entropy}}\\
\addlinespace[1pt]
KL coef $\beta_{\mathrm{KL}}$ & $0.001$ \\
KL type & \texttt{low\_var\_kl} \\
Entropy coefficient & $0$ \\
\midrule

\multicolumn{2}{@{}l}{\textbf{FCO (cross-state)}}\\
\addlinespace[1pt]
FCO weight $\lambda_{\mathrm{FCO}}$ & $0.01$ \\
FCO inverse temperature $\tau$ & $0.005$ \\
\midrule

\multicolumn{2}{@{}l}{\textbf{Sampling / Multi-turn Regeneration}}\\
\addlinespace[1pt]
Group size $K$ & $8$ \\
Temperature & $1.0$ \\
Max prompt length & $2048$ \\
Max response length & $8192$ \\
Max feedback length & $1024$ \\
Max regeneration turns & $2$ \\
Feedback subgroup size $K_{\mathrm{fb}}$ & $2$ \\
Feedback subgroups $g$ & $\lceil 8/2\rceil = 4$ \\

\bottomrule
\end{tabularx}
\caption{Key hyperparameters for MulFeRL.}
\label{tab:rl_details}
\end{table}

\begin{table}[t]
\centering
\small
\setlength{\tabcolsep}{7pt}
\renewcommand{\arraystretch}{1.10}
\begin{tabularx}{\linewidth}{@{}X r@{}}
\toprule
\textbf{Parameter} & \textbf{Value} \\
\midrule
Feedback provider & GPT-4o \\
Initial rollout group size $K$ & $8$ \\
Temperature & $1.0$ \\
Max prompt length & $2048$ \\
Max response length & $8192$ \\
Max feedback length & $1024$ \\
Total RL steps & $400$ \\
RL reward & final-answer correctness \\
Feedback-modeling coefficient $\lambda_{\mathrm{FM}}$ & $0.2$ \\
Feedback-modeling target & GPT-4o critique tokens \\
Structured MulFeRL format & no \\
FCO objective & no \\
Test-time external feedback & disabled by default \\
\bottomrule
\end{tabularx}
\caption{Additional hyperparameters for RLTF-FM. Shared RL optimization and sampling settings follow Table~\ref{tab:rl_details}.}
\label{tab:rltf_fm_details}
\end{table}

\subsubsection{Self-Distillation Details}
\label{app:hparams:self_distill}

We implement SDPO~\citep{hubotter2026reinforcement} as a self-distillation baseline with the same rollout and feedback budget as the RL baselines. SDPO converts textual feedback into dense token-level supervision by comparing the no-feedback policy with a feedback-conditioned teacher distribution. In the original formulation, SDPO treats a feedback-conditioned policy as a self-teacher and distills its feedback-informed next-token predictions back into the no-feedback policy; this provides token-level credit assignment without requiring an external reward model or an additional refinement rollout.

In our implementation, we maintain an EMA-updated copy of the student as the feedback-conditioned teacher. Concretely, for each training prompt $x$, the trainable student policy first samples an initial response $y$ under the no-feedback context. We then query GPT-4o to obtain textual feedback $c$ for the sampled response. The EMA teacher is conditioned on the feedback-augmented context $(x,y,c)$ and re-scores the same response tokens in $y$; gradients are stopped through this teacher distribution. The trainable student is optimized under the original no-feedback context to match the feedback-conditioned teacher distribution on the sampled response tokens. After each optimization step, the teacher parameters are updated as $\theta_{\mathrm{tea}}\leftarrow \mu\theta_{\mathrm{tea}}+(1-\mu)\theta_{\mathrm{stu}}$.

SDPO uses the same training prompts, rollout group size, sampling temperature, maximum prompt length, maximum response length, maximum feedback length, and total training steps as the other feedback-based baselines. Unlike MulFeRL, SDPO does not perform feedback-guided regeneration, does not use the structured MulFeRL output format, and does not use the FCO objective. At test time, SDPO is evaluated in the same single-pass setting without external feedback unless explicitly stated otherwise.

\begin{table}[t]
\centering
\small
\setlength{\tabcolsep}{7pt}
\renewcommand{\arraystretch}{1.10}
\begin{tabularx}{\linewidth}{@{}X r@{}}
\toprule
\textbf{Parameter} & \textbf{Value} \\
\midrule
Feedback provider & GPT-4o \\
Student initialization & backbone checkpoint \\
Teacher initialization & backbone checkpoint \\
Teacher update & EMA after each optimizer step \\
EMA decay $\mu$ & $0.999$ \\
Initial rollout group size $K$ & $8$ \\
Temperature & $1.0$ \\
Max prompt length & $2048$ \\
Max response length & $8192$ \\
Max feedback length & $1024$ \\
Total training steps & $400$ \\
Distillation target & feedback-conditioned teacher logits \\
Teacher context & prompt + sampled response + feedback \\
Student context & prompt only \\
Extra refinement sampling & no \\
Structured MulFeRL format & no \\
FCO objective & no \\
Test-time external feedback & disabled by default \\
\bottomrule
\end{tabularx}
\caption{Key hyperparameters for SDPO. We use an EMA-updated copy of the student as the feedback-conditioned teacher. Shared sampling and length settings follow Table~\ref{tab:rl_details}.}
\label{tab:sdpo_details}
\end{table}

\subsubsection{SFT-Training Details}
\label{app:hparams:sft}
We include several supervised finetuning baselines (e.g., SFT, RAFT, and CITL-FT), which are trained with a standard supervised objective.
Table~\ref{tab:sft_details} summarizes the key hyperparameters used for all SFT-style runs. Figure~\ref{fig:sft_ini}, Figure~\ref{fig:citl_critique}, and Figure~\ref{fig:citl_refinement} summarize the prompts used for the supervised finetuning baselines. 
\begin{table}[t]
\centering
\small
\setlength{\tabcolsep}{7pt}
\renewcommand{\arraystretch}{1.10}
\begin{tabularx}{\linewidth}{@{}X r@{}}
\toprule
\textbf{Parameter} & \textbf{Value} \\
\midrule
Max Prompt Length & $2048$ \\
Max Response Length & $8192$ \\
Per-device micro-batch & $4$ \\
Learning rate & $1\times 10^{-5}$ \\
Total steps & $600$ \\
\bottomrule
\end{tabularx}
\caption{Key hyperparameters for SFT training (if applicable).}
\label{tab:sft_details}
\end{table}

\subsection{Evaluation Details}
\label{app:hparams:eval}

We evaluate with a decoding temperature of $1.0$ and keep the maximum input/output lengths identical to those used during training. For each dataset, we report the mean performance over 5 independent runs.

\subsubsection{Shared Inference Setup}
\label{app:shared_infer}

\paragraph{Backbones and feedback simulator.}
We use Qwen3-4B-Instruct-2507 \citep{qwen3technicalreport} and Qwen2.5-7B-Base \citep{DBLP:journals/corr/abs-2412-15115} as backbones.
When external feedback is enabled, GPT-4o \citep{openai_gpt4o_system_card_2024} serves as the feedback simulator $\psi$ unless otherwise stated.

\paragraph{Decoding and length limits.}
We use a unified decoding configuration across methods for fair comparison.
We cap the maximum generation length to 8k tokens and keep the stopping criteria consistent across all evaluated models.

\paragraph{Scoring and evaluation.}
We report Pass@1 using a single model output per instance, and report results using the best checkpoint selected on the validation set for each method.
Unless explicitly stated, evaluation is \emph{single-pass} and does not use any external feedback at test time.
We use each benchmark’s official or commonly adopted evaluation scripts/format checks for verification.
Unless explicitly stated, for MulFeRL we enforce the fixed output format shown in Figure~\ref{fig:output_format}: when feedback is available, it is explicitly placed inside a dedicated \texttt{<feedback>} tag within the \texttt{<thinking>} block; when feedback is absent, the model may either generate self feedback in the same tag or leave it empty (depending on the model’s behavior). This structured format is used to remain consistent with the output convention during training. For other RL methods, we follow their default generation interface and evaluate them using the benchmark-provided evaluation scripts.


\subsubsection{Main Results (Table~\ref{tab:main})}
\label{app:main_eval}

All methods are evaluated in the single-pass setting described in  \S~\ref{app:shared_infer}.
No external feedback is used at test time.
We keep decoding, maximum length identical across methods.
For all methods, we employ the benchmark-provided (or commonly used) answer verification scripts; in addition, MulFeRL applies a training-consistent formatting constraint.

\subsubsection{Ablation Study (Table~\ref{tab:ablation-MulFeRL-math})}
\label{app:ablation_eval}

We evaluate all ablated variants under the same single-pass setup.
The ablations differ only in the algorithmic components enabled during finetuning:
\begin{itemize}[leftmargin=*, itemsep=0.25em, topsep=0.25em]
    \item \textbf{w/o Regeneration:} disables multi-turn regeneration; the learned policy is evaluated as usual in a single pass.
    \item \textbf{w/o FCO:} removes the cross-state FCO branch; evaluation remains single-pass.
    \item \textbf{w/o Feedback injection:} replaces structured injection with a plain prompting variant that provides the same feedback content; evaluation remains single-pass.
\end{itemize}

\subsubsection{Impact of Feedback Simulators (Table~\ref{tab:simulator_qwen3})}
\label{app:simulator_eval}

For all methods, we study the effect of the feedback simulator by changing only $\psi$ during training while keeping the remaining training and evaluation settings fixed.
At evaluation time, we use the same single-pass protocol without external feedback, so the reported differences reflect how simulator choice affects the learned policy.

\subsubsection{Test-time Multi-turn Feedback (Figure~\ref{fig:testtime_multiturn})}
\label{app:testtime_eval}

To evaluate test-time feedback utilization, we enable multi-turn interaction during inference. For MulFeRL, we query $\psi$ for feedback using the template in Figure~\ref{fig:feedback_system_prompt} (single rollout; no merging), and perform feedback-conditioned regeneration using the prompt in Figure~\ref{fig:feedback_injection_regen_prompt}. For GRPO and Critique-GRPO, we follow the prompting setup in previous work \citep{zhang2025critique} for both feedback elicitation and critique-guided regeneration. In all cases, feedback is constrained to identify errors and suggest repair steps without revealing the correct answer. We report Pass@1 after each revision turn.

\subsubsection{Training Efficiency Measurement}
\label{sec:appendix_efficiency}
All methods are trained under the same hardware/software stack and data pipeline described in the main implementation details.

\subsubsection{Impact of Multi-turn Training (Turn Budget Ablation)}
We vary the per-step regeneration turn budget $T_{\text{train}} \in \{0,1,2,3,4,5\}$, while keeping other hyperparameters fixed.


\begin{table*}[t]
\centering
\footnotesize
\setlength{\tabcolsep}{5pt}
\renewcommand{\arraystretch}{1.06}
\begin{tabular}{@{}l l p{0.56\textwidth}@{}}
\toprule
\textbf{Domain} & \textbf{Dataset} & \textbf{Link} \\
\midrule
Training & OpenR1-Math-220k &
\url{https://huggingface.co/datasets/open-r1/OpenR1-Math-220k} \\
\midrule
Math eval & MATH-500 &
\url{https://huggingface.co/datasets/HuggingFaceH4/MATH-500} \\
Math eval & Minerva-Math &
\url{https://huggingface.co/datasets/math-ai/minervamath} \\
Math eval & OlympiadBench &
\url{https://huggingface.co/datasets/Hothan/OlympiadBench} \\
Math eval & AIME 2024 &
\url{https://huggingface.co/datasets/HuggingFaceH4/aime_2024} \\
Math eval & AMC 2023  &
\url{https://huggingface.co/datasets/AI-MO/aimo-validation-amc} \\
\midrule
Science \& General eval & MMLU-Pro &
\url{https://huggingface.co/datasets/TIGER-Lab/MMLU-Pro} \\
Science \& General eval & TheoremQA &
\url{https://huggingface.co/datasets/TIGER-Lab/TheoremQA} \\
Science \& General eval & GPQA-Diamond  &
\url{https://huggingface.co/datasets/Idavidrein/gpqa} \\
\bottomrule
\end{tabular}
\caption{Datasets used in our experiments.}
\label{tab:datasets}
\end{table*}

\section{Datasets Details}
\label{app:datasets}
Table~\ref{tab:datasets} summarizes the datasets used in our experiments and provides the corresponding source links.

\section{Additional Experiments}
\label{app:extra}

We organize the additional experiments around four questions: whether MulFeRL's gains can be explained by formatting or teacher access, which components and feedback sources matter, how robust the method is to imperfect feedback, and whether the gains persist under larger compute and across independent runs. Unless otherwise stated, we follow the evaluation protocol in Appendix~\ref{app:hparams:eval}.

\subsection{Run-to-run Stability}
\label{app:run_to_run_stability}
Table~\ref{tab:full_std_5runs} reports mean and standard deviation over independent runs for the main comparison. The improvements of MulFeRL are consistent across runs and across both backbones, rather than being driven by a small number of favorable seeds. The standard deviations are comparable to the surrounding baselines, while the average performance remains higher on the main in-domain and out-of-domain evaluations. This supports the statistical stability of the reported gains.

\begin{table*}[t]
\centering
\scriptsize
\setlength{\tabcolsep}{3.2pt}
\renewcommand{\arraystretch}{1.10}

\begin{tabular}{@{}lccccccccc@{}}
\toprule
\multirow{2}{*}{\textbf{Method}} &
\multicolumn{5}{c}{\textbf{Math (ID)}} &
\multicolumn{3}{c}{\textbf{Science \& General (OOD)}} &
\multirow{2}{*}{\textbf{Avg.}} \\
\cmidrule(lr){2-6}\cmidrule(lr){7-9}
& \textbf{AMC23}
& \textbf{AIME24}
& \makecell{\textbf{Olympiad}\\\textbf{Bench}}
& \makecell{\textbf{MATH}\\\textbf{500}}
& \makecell{\textbf{Minerva}\\\textbf{Math}}
& \makecell{\textbf{MMLU}\\\textbf{Pro}}
& \makecell{\textbf{GPQA}\\\textbf{Diamond}}
& \makecell{\textbf{Theorem}\\\textbf{QA}}
& \\
\midrule

\multicolumn{10}{@{}l}{\textit{Qwen2.5-7B-Base}}\\
Base Model
& 34.50 $\pm$ 1.12 & 13.33 $\pm$ 2.36 & 28.13 $\pm$ 0.94 & 55.24 $\pm$ 0.67 & 18.31 $\pm$ 1.25
& 45.06 $\pm$ 0.76 & 27.68 $\pm$ 1.03 & 19.00 $\pm$ 1.16 & 30.16 $\pm$ 0.76 \\
SFT
& 38.50 $\pm$ 1.37 & 12.00 $\pm$ 1.83 & 27.27 $\pm$ 1.09 & 57.12 $\pm$ 0.50 & 21.03 $\pm$ 0.89
& 47.20 $\pm$ 0.91 & 28.08 $\pm$ 0.78 & 24.10 $\pm$ 1.12 & 31.91 $\pm$ 0.60 \\
RAFT
& 46.50 $\pm$ 1.37 & 9.33 $\pm$ 1.49 & 29.58 $\pm$ 1.08 & 61.92 $\pm$ 0.73 & 17.50 $\pm$ 1.30
& 46.06 $\pm$ 0.79 & 23.94 $\pm$ 1.09 & 21.90 $\pm$ 0.96 & 32.09 $\pm$ 0.78 \\
CITL-FT
& 41.50 $\pm$ 1.37 & 15.33 $\pm$ 1.83 & 30.89 $\pm$ 0.87 & 63.00 $\pm$ 0.54 & 19.34 $\pm$ 1.11
& 47.86 $\pm$ 0.77 & 27.07 $\pm$ 0.85 & 23.65 $\pm$ 1.00 & 33.58 $\pm$ 0.62 \\
SDPO
& 41.50 $\pm$ 1.37 & 14.67 $\pm$ 1.83 & 35.91 $\pm$ 0.98 & 69.72 $\pm$ 0.72 & 28.24 $\pm$ 1.08
& 50.90 $\pm$ 0.80 & 32.53 $\pm$ 0.97 & 37.20 $\pm$ 1.14 & 38.83 $\pm$ 0.66 \\
GRPO
& 42.00 $\pm$ 1.12 & 16.00 $\pm$ 1.49 & 36.77 $\pm$ 1.04 & 70.84 $\pm$ 0.76 & 28.97 $\pm$ 0.87
& 51.10 $\pm$ 0.66 & 33.43 $\pm$ 1.03 & 37.55 $\pm$ 1.19 & 39.58 $\pm$ 0.68 \\
Dr.GRPO
& 41.00 $\pm$ 1.37 & 14.67 $\pm$ 1.83 & 35.79 $\pm$ 0.80 & 73.20 $\pm$ 0.50 & 30.37 $\pm$ 1.16
& 51.88 $\pm$ 0.93 & 33.13 $\pm$ 0.84 & 40.08 $\pm$ 0.87 & 40.02 $\pm$ 0.64 \\
Critique-GRPO
& 45.50 $\pm$ 1.12 & 20.67 $\pm$ 1.49 & 38.64 $\pm$ 0.97 & 74.96 $\pm$ 0.63 & 33.82 $\pm$ 0.82
& 52.34 $\pm$ 0.72 & 36.97 $\pm$ 0.97 & 39.75 $\pm$ 1.08 & 42.83 $\pm$ 0.60 \\
RLTF-FM
& 42.00 $\pm$ 1.12 & 15.33 $\pm$ 1.83 & 36.05 $\pm$ 0.95 & 69.88 $\pm$ 0.70 & 28.46 $\pm$ 1.02
& 50.76 $\pm$ 0.78 & 32.93 $\pm$ 0.96 & 37.65 $\pm$ 1.13 & 39.13 $\pm$ 0.65 \\
MulFeRL
& \textbf{50.00 $\pm$ 1.77} & \textbf{24.00 $\pm$ 1.49} & \textbf{42.49 $\pm$ 0.82} & \textbf{78.72 $\pm$ 0.55} & \textbf{37.87 $\pm$ 1.08}
& \textbf{54.89 $\pm$ 0.78} & \textbf{38.99 $\pm$ 0.84} & \textbf{43.08 $\pm$ 1.03} & \textbf{46.26 $\pm$ 0.65} \\

\midrule

\multicolumn{10}{@{}l}{\textit{Qwen3-4B-Inst}}\\
Base Model
& 59.00 $\pm$ 1.37 & 46.00 $\pm$ 1.49 & 45.13 $\pm$ 0.97 & 76.68 $\pm$ 0.57 & 42.43 $\pm$ 0.75
& 58.56 $\pm$ 0.76 & 35.05 $\pm$ 0.81 & 39.85 $\pm$ 0.94 & 50.34 $\pm$ 0.60 \\
SFT
& 62.50 $\pm$ 1.77 & 47.33 $\pm$ 1.49 & 49.17 $\pm$ 0.93 & 78.12 $\pm$ 0.45 & 46.32 $\pm$ 0.98
& 60.46 $\pm$ 0.72 & 36.26 $\pm$ 0.60 & 41.93 $\pm$ 0.79 & 52.76 $\pm$ 0.54 \\
RAFT
& 62.00 $\pm$ 1.12 & 46.00 $\pm$ 1.49 & 48.34 $\pm$ 1.08 & 78.68 $\pm$ 0.50 & 45.29 $\pm$ 0.72
& 59.53 $\pm$ 0.57 & 37.47 $\pm$ 0.85 & 41.05 $\pm$ 0.91 & 52.30 $\pm$ 0.58 \\
CITL-FT
& 62.00 $\pm$ 1.12 & 48.00 $\pm$ 1.83 & 48.55 $\pm$ 0.77 & 79.64 $\pm$ 0.53 & 47.13 $\pm$ 0.96
& 60.18 $\pm$ 0.68 & 36.87 $\pm$ 0.78 & 42.65 $\pm$ 0.67 & 53.13 $\pm$ 0.55 \\
SDPO
& 79.50 $\pm$ 1.12 & 58.00 $\pm$ 1.83 & 59.55 $\pm$ 0.86 & 87.80 $\pm$ 0.47 & 50.07 $\pm$ 0.98
& 63.00 $\pm$ 0.72 & 43.84 $\pm$ 0.80 & 50.30 $\pm$ 0.88 & 61.51 $\pm$ 0.56 \\
GRPO
& 78.50 $\pm$ 1.37 & 57.33 $\pm$ 1.49 & 59.05 $\pm$ 0.89 & 87.84 $\pm$ 0.52 & 51.25 $\pm$ 1.03
& 62.18 $\pm$ 0.55 & 43.94 $\pm$ 0.72 & 49.85 $\pm$ 0.99 & 61.24 $\pm$ 0.62 \\
Dr.GRPO
& 79.00 $\pm$ 1.37 & 56.67 $\pm$ 2.36 & 60.36 $\pm$ 0.71 & 88.28 $\pm$ 0.34 & 49.19 $\pm$ 1.12
& 63.33 $\pm$ 0.84 & 42.83 $\pm$ 0.95 & 50.68 $\pm$ 0.76 & 61.29 $\pm$ 0.57 \\
Critique-GRPO
& 84.00 $\pm$ 1.37 & 62.67 $\pm$ 1.49 & 62.40 $\pm$ 0.79 & 88.64 $\pm$ 0.49 & 50.29 $\pm$ 0.77
& 64.67 $\pm$ 0.67 & 45.45 $\pm$ 0.69 & 51.05 $\pm$ 0.95 & 63.65 $\pm$ 0.54 \\
RLTF-FM
& 82.50 $\pm$ 1.77 & 60.67 $\pm$ 1.49 & 61.54 $\pm$ 0.82 & 88.12 $\pm$ 0.46 & 49.63 $\pm$ 0.90
& 64.25 $\pm$ 0.70 & 44.65 $\pm$ 0.75 & 50.85 $\pm$ 0.91 & 62.78 $\pm$ 0.55 \\
MulFeRL
& \textbf{90.50 $\pm$ 1.12} & \textbf{68.00 $\pm$ 1.83} & \textbf{68.13 $\pm$ 0.73} & \textbf{90.24 $\pm$ 0.43} & \textbf{55.96 $\pm$ 0.95}
& \textbf{68.08 $\pm$ 0.69} & \textbf{50.20 $\pm$ 0.80} & \textbf{55.75 $\pm$ 0.93} & \textbf{68.36 $\pm$ 0.59} \\

\bottomrule
\end{tabular}
\caption{Results reported as mean $\pm$ standard deviation over 5 independent runs, where each run is evaluated at its best validation checkpoint.}
\label{tab:full_std_5runs}

\end{table*}

\subsection{Controlled Baseline Comparisons}
\label{app:extra:controlled}

\subsubsection{Unified Output Format}
\label{app:output_format}
To control for confounding effects caused by differences in output formatting, we align the prompts and output constraints of GRPO and Critique-GRPO to exactly match those used by MulFeRL. Specifically, both baselines use the same input prompt (Figure~\ref{fig:input_pem}) as MulFeRL; for Critique-GRPO, the prompts for getting feedback (Figure~\ref{fig:feedback_system_prompt}) and for feedback injection (Figure~\ref{fig:feedback_injection_regen_prompt}) are also identical to MulFeRL. Finally, we enforce the same structured output-format constraint as in Figure~\ref{fig:output_format} for all methods. The corresponding results are reported in Table~\ref{tab:main_rl_only}.

Under this controlled setting, GRPO shows a slight performance drop compared to the main-table results. A plausible explanation is that the structured format introduces additional generation overhead (consuming token budget) and partially limits the model’s freedom to organize reasoning and intermediate steps, reducing solution flexibility on harder instances. In addition, strict formatting can increase the risk of format violations or truncation, which lowers the fraction of valid outputs.

Critique-GRPO also underperforms its original numbers. Although it employs feedback injection, the amount of injectable feedback is relatively limited in this setting, preventing the model from fully benefiting from critique signals. Meanwhile, the structured template and strict formatting further constrain how the model organizes reasoning and intermediate steps, and can increase the likelihood of format violations/truncation. Together, these effects can shrink the effective solution space on complex problems and reduce the valid-output rate, thereby offsetting and potentially outweighing the gains from feedback. Overall, this alignment study suggests that a unified output format alone does not account for our improvements; MulFeRL’s gains are more fundamentally driven by feedback-guided iterative refinement and its associated optimization objectives.

\providecommand{\pred}[1]{#1$^\dagger$}

\begin{table*}[t]
\centering
\small
\setlength{\tabcolsep}{5pt}
\renewcommand{\arraystretch}{1.12}

\begin{tabular}{@{}lccccccccc@{}}
\toprule
\multirow{2}{*}{\textbf{Method}} &
\multicolumn{5}{c}{\textbf{Math (ID)}} &
\multicolumn{3}{c}{\textbf{Science \& General (OOD)}} &
\multirow{2}{*}{\textbf{Avg.}} \\
\cmidrule(lr){2-6}\cmidrule(lr){7-9}
& \textbf{AMC23}
& \textbf{AIME24}
& \makecell{\textbf{Olympiad}\\\textbf{Bench}}
& \makecell{\textbf{MATH}\\\textbf{500}}
& \makecell{\textbf{Minerva}\\\textbf{MATH}}
& \makecell{\textbf{MMLU}\\\textbf{Pro}}
& \makecell{\textbf{GPQA}\\\textbf{Diamond}}
& \makecell{\textbf{Theorem}\\\textbf{QA}}
& \\
\midrule

\multicolumn{10}{@{}l}{\textit{Qwen2.5-7B-Base}}\\
GRPO
& 39.50 & 16.00 & 33.68 & 67.40 & 26.32 & 48.40 & 30.20 & 34.60 & 37.01 \\
SDPO
& 38.00 & 14.00 & 34.51 & 66.72 & 26.69
& 47.80 & 30.00 & 33.95 & 36.46 \\
Critique-GRPO
& 43.50 & 20.67 & 36.23 & 72.44 & 31.47 & 50.09 & 34.44 & 37.45 & 40.79 \\
RLTF-FM
& 38.50 & 14.67 & 34.75 & 66.80 & 26.25
& 48.10 & 30.51 & 34.35 & 36.74 \\
MulFeRL
& \textbf{50.00} & \textbf{24.00} & \textbf{42.49} & \textbf{78.72} & \textbf{37.87}
& \textbf{54.89} & \textbf{38.99} & \textbf{43.08} & \textbf{46.26} \\

\midrule

\multicolumn{10}{@{}l}{\textit{Qwen3-4B-Inst}}\\
GRPO
& 75.50 & 57.33 & 55.58 & 84.72 & 48.53 & 59.13 & 41.31 & 46.15 & 58.53 \\
SDPO
& 76.50 & 57.33 & 55.99 & 84.20 & 47.50
& 59.60 & 40.61 & 46.80 & 58.57 \\
Critique-GRPO
& 82.00 & 62.67 & 59.53 & 86.20 & 48.01 & 61.92 & 43.03 & 48.50 & 61.48 \\
RLTF-FM
& 79.00 & 59.33 & 57.63 & 85.36 & 47.57
& 60.98 & 42.32 & 47.95 & 60.02 \\
MulFeRL
& \textbf{90.50} & \textbf{68.00} & \textbf{68.13} & \textbf{90.24} & \textbf{55.96}
& \textbf{68.08} & \textbf{50.20} & \textbf{55.75} & \textbf{68.36} \\

\bottomrule
\end{tabular}
\caption{Evaluation results (Pass@1) on mathematical reasoning (ID) and scientific/general reasoning (OOD) benchmarks under a unified MulFeRL-style feedback-injection output format across all methods.}
\label{tab:main_rl_only}

\end{table*}

\subsubsection{Same-budget GPT-4o Distillation}
\label{app:distill_vs_mulferl}
Table~\ref{tab:distill_vs_mulferl_same_api} compares MulFeRL against a GPT-4o distillation baseline under approximately matched training-time GPT-4o API cost. Across both backbones, MulFeRL is consistently stronger than the distillation baseline on both in-domain mathematical reasoning and out-of-domain reasoning benchmarks. This suggests that the improvement is not simply due to querying GPT-4o or exposing the student to stronger model outputs. Instead, the benefit comes from how MulFeRL uses external feedback: the model diagnoses and revises its own failed reasoning, and the verifier then turns successful feedback-guided revisions into RL supervision. In contrast, distillation mainly encourages imitation of teacher-generated outputs, which can be less aligned with preserving and improving the student's own reasoning behavior.

\begin{table*}[t]
\centering
\small
\setlength{\tabcolsep}{5pt}
\renewcommand{\arraystretch}{1.12}
\begin{tabular}{@{}lccccccccc@{}}
\toprule
\multirow{2}{*}{\textbf{Method}} &
\multicolumn{5}{c}{\textbf{Math (ID)}} &
\multicolumn{3}{c}{\textbf{Science \& General (OOD)}} &
\multirow{2}{*}{\textbf{Avg.}} \\
\cmidrule(lr){2-6}\cmidrule(lr){7-9}
& \textbf{AMC23}
& \textbf{AIME24}
& \makecell{\textbf{Olympiad}\\\textbf{Bench}}
& \makecell{\textbf{MATH}\\\textbf{500}}
& \makecell{\textbf{Minerva}\\\textbf{MATH}}
& \makecell{\textbf{MMLU}\\\textbf{Pro}}
& \makecell{\textbf{GPQA}\\\textbf{Diamond}}
& \makecell{\textbf{Theorem}\\\textbf{QA}}
& \\
\midrule

\multicolumn{10}{@{}l}{\textit{Qwen2.5-7B-Base}}\\
GPT-4o Distillation
& 47.61 & 21.33 & 40.52 & 77.58 & 35.27 & 52.71 & 37.11 & 40.76 & 44.11 \\
MulFeRL
& \textbf{50.00} & \textbf{24.00} & \textbf{42.49} & \textbf{78.72} & \textbf{37.87}
& \textbf{54.89} & \textbf{38.99} & \textbf{43.08} & \textbf{46.26} \\

\midrule

\multicolumn{10}{@{}l}{\textit{Qwen3-4B-Inst}}\\
GPT-4o Distillation
& 88.18 & 63.67 & 65.74 & 89.33 & 53.76 & 66.49 & 48.88 & 53.71 & 66.22 \\
MulFeRL
& \textbf{90.50} & \textbf{68.00} & \textbf{68.13} & \textbf{90.24} & \textbf{55.96}
& \textbf{68.08} & \textbf{50.20} & \textbf{55.75} & \textbf{68.36} \\

\bottomrule
\end{tabular}
\caption{Comparison between GPT-4o distillation and MulFeRL when the distillation baseline is constrained to use approximately the same \emph{training-time GPT-4o API cost in USD} as MulFeRL. All results are reported as 5-run means, where each run is evaluated at its best validation checkpoint.}
\label{tab:distill_vs_mulferl_same_api}
\end{table*}

\subsection{Component and Simulator Analyses}
\label{app:extra:components}

\subsubsection{Component Ablations}
\label{app:ablation}
The complete ablation results are reported in Table~\ref{tab:ablation_full}. The conclusions are consistent with the main experiments. Performance degrades across all ablations, indicating that each component contributes materially to MulFeRL. Disabling multi-turn regeneration yields the largest drop, consistent with the observation that single-turn scalar rewards provide a weak learning signal on failed samples. Removing the FCO branch while keeping regeneration also leads to a clear decline, suggesting that regeneration alone does not reliably convert feedback-induced solved-state transitions into a stable training signal: when feedback moves an all-failed group to an all-solved group, within-state GRPO has no reward contrast, and cross-state FCO is needed to consolidate the improvement direction. Finally, ablating feedback injection consistently hurts performance, implying that structured injection conditions the model's reasoning on feedback more faithfully than plain prompting, improving feedback utilization and stabilizing training. We omit Qwen2.5-7B-Base as a simulator because, without instruction tuning, it frequently violates the required output schema, making feedback extraction difficult.

\begin{table*}[t]
  \centering
  \small
  \setlength{\tabcolsep}{4.6pt}
  \renewcommand{\arraystretch}{1.12}
  \begin{tabular}{@{}lccccccccc@{}}
    \toprule
    \multirow{2}{*}{\textbf{Method}} &
    \multicolumn{5}{c}{\textbf{Math (ID)}} &
    \multicolumn{3}{c}{\textbf{Science \& General (OOD)}} &
    \multirow{2}{*}{\textbf{Avg.}} \\
    \cmidrule(lr){2-6}\cmidrule(lr){7-9}
    & \textbf{AMC23}
    & \textbf{AIME24}
    & \makecell{\textbf{Olympiad}\\\textbf{Bench}}
    & \makecell{\textbf{MATH}\\\textbf{500}}
    & \makecell{\textbf{Minerva}\\\textbf{MATH}}
    & \makecell{\textbf{MMLU}\\\textbf{Pro}}
    & \makecell{\textbf{GPQA}\\\textbf{Diamond}}
    & \makecell{\textbf{Theorem}\\\textbf{QA}}
    & \\
    \midrule

    \multicolumn{10}{@{}l}{\textit{Qwen2.5-7B-Base}}\\
    MulFeRL
    & \textbf{50.00} & \textbf{24.00} & \textbf{42.49} & \textbf{78.72} & \textbf{37.87}
    & \textbf{54.89} & \textbf{38.99} & \textbf{43.08} & \textbf{46.26} \\
    w/o Regeneration
    & 42.20 & 16.00 & 36.77 & 70.84 & 28.97 & 51.10 & 33.43 & 37.55 & 39.61 \\
    w/o FCO
    & 46.20 & 19.67 & 39.38 & 76.12 & 32.13 & 52.24 & 35.15 & 39.45 & 42.54 \\
    w/o Feedback injection
    & 47.10 & 20.67 & 40.39 & 77.68 & 34.34 & 53.09 & 36.87 & 41.18 & 43.92 \\

    \midrule

    \multicolumn{10}{@{}l}{\textit{Qwen3-4B-Inst}}\\
    MulFeRL
    & \textbf{90.50} & \textbf{68.00} & \textbf{68.13} & \textbf{90.24} & \textbf{55.96}
    & \textbf{68.08} & \textbf{50.20} & \textbf{55.75} & \textbf{68.36} \\
    w/o Regeneration
    & 78.60 & 57.33 & 59.05 & 87.84 & 51.25 & 62.18 & 43.94 & 49.85 & 61.26 \\
    w/o FCO
    & 87.70 & 65.67 & 65.79 & 88.08 & 53.16 & 66.36 & 47.68 & 53.20 & 65.96 \\
    w/o Feedback injection
    & 88.50 & 66.33 & 66.62 & 88.76 & 54.04 & 66.97 & 48.79 & 54.10 & 66.76 \\

    \bottomrule
  \end{tabular}
  \caption{Ablation of MulFeRL components on Qwen3-4B-Inst and Qwen2.5-7B-Base. Numbers are Pass@1 (\%).}
  \label{tab:ablation_full}
\end{table*}

\subsubsection{Feedback Simulators}
\label{sub:simulator}
The complete results are given in Table~\ref{tab:simulators_full_qwen3}. The same takeaway as in the main paper holds: simulator quality strongly correlates with the final performance of MulFeRL. As shown in Table~\ref{tab:simulators_full_qwen3}, replacing the feedback source with a more capable simulator consistently improves downstream accuracy. Notably, even using the training backbone itself as the simulator is competitive and surpasses a weaker small model, indicating that MulFeRL can effectively leverage reasonably good self-generated critiques and is not tied to a specific simulator. Moving to stronger external simulators yields additional gains: GPT-4o-mini provides a clear improvement over self-feedback, and GPT-4o further increases performance. A plausible explanation is that higher-quality simulators produce more precise failure localization and more targeted, executable suggestions, which in turn guide regeneration toward better candidates and provide cleaner learning signals, leading to better generalization on difficult mathematical reasoning benchmarks.

\begin{table*}[t]
  \centering
  \footnotesize
  \setlength{\tabcolsep}{3.5pt}
  \renewcommand{\arraystretch}{1.12}

  \begin{tabular}{@{}lccccccccc@{}}
    \toprule
    \multirow{2}{*}{\textbf{Feedback simulator}} &
    \multicolumn{5}{c}{\textbf{Math (ID)}} &
    \multicolumn{3}{c}{\textbf{Science \& General (OOD)}} &
    \multirow{2}{*}{\textbf{Avg.}} \\
    \cmidrule(lr){2-6}\cmidrule(lr){7-9}
    & \textbf{AMC23}
    & \textbf{AIME24}
    & \makecell{\textbf{Olympiad}\\\textbf{Bench}}
    & \makecell{\textbf{MATH}\\\textbf{500}}
    & \makecell{\textbf{Minerva}\\\textbf{Math}}
    & \makecell{\textbf{MMLU}\\\textbf{Pro}}
    & \makecell{\textbf{GPQA}\\\textbf{Diamond}}
    & \makecell{\textbf{Theorem}\\\textbf{QA}}
    & \\
    \midrule

    \multicolumn{10}{@{}l@{}}{\textit{Backbone: Qwen2.5-7B-Base}}\\
    Qwen3-1.7B
    & 47.10 & 20.00 & 38.78 & 76.92 & 33.75 & 53.09 & 36.06 & 41.10 & 43.35 \\
    GPT-4o-mini
    & 49.20 & 21.67 & 41.57 & 78.08 & 36.18 & 54.31 & 37.78 & 42.25 & 45.13 \\
    GPT-4o
    & \textbf{50.00} & \textbf{24.00} & \textbf{42.49} & \textbf{78.72} & \textbf{37.87}
    & \textbf{54.89} & \textbf{38.99} & \textbf{43.08} & \textbf{46.26} \\

    \midrule

    \multicolumn{10}{@{}l@{}}{\textit{Backbone: Qwen3-4B-Inst}}\\
    Qwen3-1.7B
    & 84.00 & 58.67 & 58.81 & 86.00 & 48.75 & 62.50 & 43.43 & 48.40 & 61.32 \\
    Self (base checkpoint)
    & 88.40 & 64.00 & 63.62 & 88.68 & 51.84 & 67.18 & 46.98 & 54.26 & 65.62 \\
    GPT-4o-mini
    & 89.10 & 66.33 & 66.94 & 89.24 & 52.65 & 67.32 & 48.59 & 54.55 & 66.84 \\
    GPT-4o
    & \textbf{90.50} & \textbf{68.00} & \textbf{68.13} & \textbf{90.24} & \textbf{55.96}
    & \textbf{68.08} & \textbf{50.20} & \textbf{55.75} & \textbf{68.36} \\

    \bottomrule
  \end{tabular}
  \caption{MulFeRL trained with different feedback simulators on Qwen3-4B-Inst and Qwen2.5-7B-Base.}
  \label{tab:simulators_full_qwen3}
\end{table*}

\subsubsection{Matched Train-time and Test-time Feedback Variants}
\label{app:training_robustness_text}
Table~\ref{tab:training_robustness} studies matched train-time and test-time feedback variants. 
For each feedback variant, methods are trained under the same setup as in the main experiments except for the feedback type, and are evaluated with one round of feedback-guided regeneration using the same feedback type as in training.
Outcome-only feedback provides a much weaker learning and revision signal than full verbal feedback, showing that binary verdicts alone do not capture the diagnostic information needed for robust improvement. 
Full-verbal feedback is consistently strongest, and MulFeRL benefits more from it than Critique-GRPO, indicating that the MulFeRL feedback format and optimization objectives make better use of rich diagnostic feedback during training and regeneration.

For compactness, the feedback-variant tables use four short names. \textit{Outcome-only} denotes binary verdict feedback only, \textit{Wrong-verdict} denotes verbal feedback induced by a fully flipped verdict, \textit{Corrupted-diagnostics} denotes verbal feedback with fully corrupted diagnostics, and \textit{Full-verbal} denotes full issue-and-fix feedback. We use ``w/o FI'' for the ablation without structured feedback injection.

\begin{table*}[t]
\centering
\footnotesize
\setlength{\tabcolsep}{2pt}
\renewcommand{\arraystretch}{1.10}
\begin{tabularx}{\textwidth}{@{}>{\raggedright\arraybackslash}Xccccccccc@{}}
\toprule
\multirow{2}{*}{\textbf{Method}} &
\multicolumn{5}{c}{\textbf{Math (ID)}} &
\multicolumn{3}{c}{\textbf{Science \& General (OOD)}} &
\multirow{2}{*}{\makecell{\textbf{Avg.}\\\textbf{(mean $\pm$ std.)}}} \\
\cmidrule(lr){2-6}\cmidrule(lr){7-9}
& \textbf{AMC23}
& \textbf{AIME24}
& \makecell{\textbf{Olympiad}\\\textbf{Bench}}
& \makecell{\textbf{MATH}\\\textbf{500}}
& \makecell{\textbf{Minerva}\\\textbf{MATH}}
& \makecell{\textbf{MMLU}\\\textbf{Pro}}
& \makecell{\textbf{GPQA}\\\textbf{Diamond}}
& \makecell{\textbf{Theorem}\\\textbf{QA}}
& \\
\midrule

\multicolumn{10}{@{}l}{\textit{Qwen2.5-7B-Base}}\\
GRPO + Full-verbal
& 44.42 & 22.67 & 39.18 & 71.68 & 31.54 & 53.64 & 35.52 & 51.96 & 43.83 $\pm$ 0.45  \\
Critique-GRPO + Outcome-only
& 45.16 & 21.33 & 38.64 & 74.90 & 33.70 & 52.50 & 36.76 & 40.92 & 42.99 $\pm$ 0.46 \\
Critique-GRPO + Full-verbal
& 47.98 & 25.33 & 41.12 & 75.84 & 35.74 & 54.42 & 38.68 & 54.37 & 46.69 $\pm$ 0.54 \\
MulFeRL + Outcome-only
& 44.98 & 20.67 & 39.74 & 74.02 & 32.88 & 53.95 & 36.61 & 41.55 & 43.05 $\pm$ 0.47 \\
MulFeRL + Full-verbal
& \textbf{54.08} & \textbf{30.67} & \textbf{46.72} & \textbf{79.82} & \textbf{41.44}
& \textbf{58.77} & \textbf{43.81} & \textbf{58.42} & \textbf{51.72 $\pm$ 0.66} \\

\midrule

\multicolumn{10}{@{}l}{\textit{Qwen3-4B-Inst}}\\
GRPO + Full-verbal
& 81.18 & 64.67 & 62.18 & 88.88 & 53.76 & 64.86 & 46.12 & 56.41 & 64.76 $\pm$ 0.38 \\
Critique-GRPO + Outcome-only
& 83.82 & 62.67 & 62.42 & 88.64 & 50.24 & 64.62 & 45.34 & 51.46 & 63.65 $\pm$ 0.43 \\
Critique-GRPO + Full-verbal
& 86.62 & 68.00 & 65.76 & 89.48 & 52.98 & 66.84 & 48.18 & 58.52 & 67.05 $\pm$ 0.50 \\
MulFeRL + Outcome-only
& 82.34 & 62.00 & 62.88 & 89.01 & 53.74 & 64.98 & 46.92 & 53.18 & 64.38 $\pm$ 0.43 \\
MulFeRL + Full-verbal
& \textbf{93.75} & \textbf{73.33} & \textbf{71.89} & \textbf{91.09} & \textbf{58.98}
& \textbf{71.33} & \textbf{54.21} & \textbf{61.82} & \textbf{72.05 $\pm$ 0.53}\\

\bottomrule
\end{tabularx}
\caption{Compact matched train-time and test-time feedback variants. For MulFeRL and Critique-GRPO, the feedback type used during training is matched to the feedback type used during one feedback-guided regeneration round at test time. GRPO has no feedback-conditioned training and is included with full-verbal test-time feedback as an anchor. Variant names are defined in the text. Values are 5-run means, with standard deviations reported for averages.}
\label{tab:training_robustness}
\end{table*}

\subsubsection{Robustness Without Structured Feedback Injection}
\label{app:feedback_injection_robustness_text}

All methods use the same trained checkpoints as in the main experiments; only at test time do we provide one round of feedback-guided regeneration with the specified feedback variant.
Table~\ref{tab:feedback_injection_robustness} repeats the compact feedback-variant analysis for the ablation without structured feedback injection. Full-verbal feedback still provides the strongest revisions, but wrong-verdict and corrupted-diagnostics feedback cause larger degradation than clean feedback. This supports the role of structured injection: placing feedback inside the reasoning interface helps the policy use useful diagnostics more faithfully and reduces sensitivity to severely misleading feedback. The result also clarifies that robustness does not come merely from applying another test-time revision round; it depends on how feedback was integrated during training and inference.

\begin{table*}[t]
\centering
\footnotesize
\setlength{\tabcolsep}{2pt}
\renewcommand{\arraystretch}{1.10}
\begin{tabularx}{\textwidth}{@{}>{\raggedright\arraybackslash}Xccccccccc@{}}
\toprule
\multirow{2}{*}{\textbf{Method}} &
\multicolumn{5}{c}{\textbf{Math (ID)}} &
\multicolumn{3}{c}{\textbf{Science \& General (OOD)}} &
\multirow{2}{*}{\makecell{\textbf{Avg.}\\\textbf{(mean $\pm$ std.)}}}  \\
\cmidrule(lr){2-6}\cmidrule(lr){7-9}
& \textbf{AMC23}
& \textbf{AIME24}
& \makecell{\textbf{Olympiad}\\\textbf{Bench}}
& \makecell{\textbf{MATH}\\\textbf{500}}
& \makecell{\textbf{Minerva}\\\textbf{MATH}}
& \makecell{\textbf{MMLU}\\\textbf{Pro}}
& \makecell{\textbf{GPQA}\\\textbf{Diamond}}
& \makecell{\textbf{Theorem}\\\textbf{QA}}
& \\
\midrule

\multicolumn{10}{@{}l}{\textit{Qwen2.5-7B-Base}}\\
MulFeRL w/o FI + Outcome-only
& 47.28 & 21.33 & 40.58 & 76.92 & 34.36 & 53.18 & 36.98 & 46.12 & 44.59 $\pm$ 0.61 \\
MulFeRL w/o FI + Wrong-verdict
& 43.24 & 17.33 & 37.42 & 75.74 & 31.62 & 50.24 & 34.28 & 34.86 & 40.59 $\pm$ 1.04 \\
MulFeRL w/o FI + Corrupted-diagnostics
& 44.04 & 18.67 & 38.32 & 75.92 & 32.32 & 50.86 & 34.88 & 37.08 & 41.51 $\pm$ 0.97 \\
MulFeRL w/o FI + Full-verbal
& \textbf{52.18} & \textbf{24.67} & \textbf{43.96} & \textbf{79.10} & \textbf{37.02}
& \textbf{55.92} & \textbf{39.52} & \textbf{56.60} & \textbf{48.62 $\pm$ 0.66} \\

\midrule

\multicolumn{10}{@{}l}{\textit{Qwen3-4B-Inst}}\\
MulFeRL w/o FI + Outcome-only
& 88.98 & 67.33 & 66.90 & 88.94 & 54.34 & 67.14 & 49.10 & 54.88 & 67.20 $\pm$ 0.55 \\
MulFeRL w/o FI + Wrong-verdict
& 85.96 & 63.33 & 63.88 & 88.22 & 52.12 & 64.44 & 46.82 & 46.20 & 63.87 $\pm$ 1.07 \\
MulFeRL w/o FI + Corrupted-diagnostics
& 86.88 & 64.67 & 65.02 & 88.38 & 52.78 & 65.22 & 47.42 & 47.40 & 64.72 $\pm$ 0.98 \\
MulFeRL w/o FI + Full-verbal
& \textbf{92.18} & \textbf{70.67} & \textbf{69.82} & \textbf{90.68} & \textbf{56.72}
& \textbf{69.88} & \textbf{51.42} & \textbf{59.82} & \textbf{70.15 $\pm$ 0.65} \\

\bottomrule
\end{tabularx}
\caption{Compact test-time feedback variants for \textbf{MulFeRL w/o feedback injection}. We write ``w/o FI'' for removing structured feedback injection; variant names are defined in the text. All feedback-conditioned variants use one round of feedback-guided regeneration at test time. Values are 5-run means, with standard deviations reported for averages.}
\label{tab:feedback_injection_robustness}
\end{table*}
\subsubsection{Test-time Feedback Variants}
\label{app:testtime_feedback_variants}
All methods use the same trained checkpoints as in the main experiments; only at test time do we provide one round of feedback-guided regeneration with the specified feedback variant.
Table~\ref{tab:feedback_disentangle_panelB} isolates the effect of test-time feedback by keeping the trained checkpoints fixed and varying only the feedback available during one regeneration round. Full-verbal feedback gives the strongest revisions, whereas outcome-only verdicts provide much weaker guidance because they do not explain how to repair the failed solution. Wrong-verdict feedback consistently hurts, and fully corrupted diagnostics also degrade MulFeRL, confirming that feedback quality matters. Under clean full-verbal feedback, MulFeRL remains strongest across backbones and evaluation groups.

\begin{table*}[t]
\centering
\footnotesize
\setlength{\tabcolsep}{2pt}
\renewcommand{\arraystretch}{1.10}
\begin{tabularx}{\textwidth}{@{}>{\raggedright\arraybackslash}Xccccccccc@{}}
\toprule
\multirow{2}{*}{\textbf{Method}} &
\multicolumn{5}{c}{\textbf{Math (ID)}} &
\multicolumn{3}{c}{\textbf{Science \& General (OOD)}} &
\multirow{2}{*}{\makecell{\textbf{Avg.}\\\textbf{(mean $\pm$ std.)}}} \\
\cmidrule(lr){2-6}\cmidrule(lr){7-9}
& \textbf{AMC23}
& \textbf{AIME24}
& \makecell{\textbf{Olympiad}\\\textbf{Bench}}
& \makecell{\textbf{MATH}\\\textbf{500}}
& \makecell{\textbf{Minerva}\\\textbf{MATH}}
& \makecell{\textbf{MMLU}\\\textbf{Pro}}
& \makecell{\textbf{GPQA}\\\textbf{Diamond}}
& \makecell{\textbf{Theorem}\\\textbf{QA}}
& \\
\midrule

\multicolumn{10}{@{}l}{\textit{Qwen2.5-7B-Base}}\\
GRPO + Outcome-only
& 42.56 & 20.00 & 37.62 & 71.04 & 30.42 & 52.26 & 34.46 & 43.12 & 41.44 $\pm$ 0.52 \\
GRPO + Wrong-verdict
& 39.14 & 14.67 & 34.98 & 69.96 & 27.44 & 49.53 & 32.22 & 33.89 & 37.73 $\pm$ 0.79 \\
GRPO + Full-verbal
& 44.42 & 22.67 & 39.18 & 71.68 & 31.54 & 53.64 & 35.52 & 51.96 & 43.83 $\pm$ 0.45 \\
Critique-GRPO + Outcome-only
& 45.42 & 21.33 & 38.88 & 74.98 & 33.88 & 52.72 & 36.98 & 41.20 & 43.17 $\pm$ 0.48 \\
Critique-GRPO + Wrong-verdict
& 42.02 & 17.67 & 36.06 & 73.96 & 31.22 & 50.14 & 34.66 & 36.83 & 40.32 $\pm$ 0.74 \\
Critique-GRPO + Full-verbal
& 47.98 & 25.33 & 41.12 & 75.84 & 35.74 & 54.42 & 38.68 & 54.37 & 46.69 $\pm$ 0.52 \\
MulFeRL + Outcome-only
& 50.66 & 25.33 & 43.36 & 78.96 & 38.48 & 55.68 & 40.02 & 45.94 & 47.30 $\pm$ 0.54 \\
MulFeRL + Wrong-verdict
& 47.56 & 20.00 & 40.12 & 78.02 & 35.44 & 52.57 & 36.21 & 40.30 & 43.78 $\pm$ 0.90 \\
MulFeRL + Corrupted-diagnostics
& 48.46 & 21.33 & 41.04 & 78.30 & 36.28 & 53.51 & 37.29 & 42.20 & 44.80 $\pm$ 0.83 \\
MulFeRL + Full-verbal
& \textbf{54.08} & \textbf{30.67} & \textbf{46.72} & \textbf{79.82} & \textbf{41.44}
& \textbf{58.77} & \textbf{43.81} & \textbf{58.42} & \textbf{51.72 $\pm$ 0.66} \\

\midrule

\multicolumn{10}{@{}l}{\textit{Qwen3-4B-Inst}}\\
GRPO + Outcome-only
& 79.64 & 60.67 & 60.56 & 88.18 & 52.44 & 63.72 & 44.92 & 51.30 & 62.68 $\pm$ 0.48 \\
GRPO + Wrong-verdict
& 77.12 & 54.67 & 57.98 & 87.16 & 49.72 & 60.48 & 42.66 & 39.84 & 58.70 $\pm$ 0.81 \\
GRPO + Full-verbal
& 81.18 & 64.67 & 62.18 & 88.88 & 53.76 & 64.86 & 46.12 & 56.41 & 64.76 $\pm$ 0.38 \\
Critique-GRPO + Outcome-only
& 84.02 & 63.33 & 62.66 & 88.70 & 50.48 & 64.82 & 45.56 & 51.76 & 63.92 $\pm$ 0.44 \\
Critique-GRPO + Wrong-verdict
& 81.82 & 58.00 & 59.78 & 87.94 & 48.24 & 62.56 & 43.54 & 45.14 & 60.88 $\pm$ 0.78 \\
Critique-GRPO + Full-verbal
& 86.62 & 68.00 & 65.76 & 89.48 & 52.98 & 66.84 & 48.18 & 58.52 & 67.05 $\pm$ 0.48 \\
MulFeRL + Outcome-only
& 90.84 & 69.33 & 68.86 & 90.39 & 56.48 & 68.86 & 51.02 & 56.92 & 69.09 $\pm$ 0.51 \\
MulFeRL + Wrong-verdict
& 88.21 & 64.67 & 66.01 & 89.73 & 54.16 & 66.43 & 47.73 & 52.60 & 66.19 $\pm$ 0.94 \\
MulFeRL + Corrupted-diagnostics
& 89.01 & 66.00 & 66.79 & 89.99 & 54.80 & 67.19 & 48.67 & 53.84 & 67.04 $\pm$ 0.86 \\
MulFeRL + Full-verbal
& \textbf{93.75} & \textbf{73.33} & \textbf{71.89} & \textbf{91.09} & \textbf{58.98}
& \textbf{71.33} & \textbf{54.21} & \textbf{61.82} & \textbf{72.05 $\pm$ 0.53} \\

\bottomrule
\end{tabularx}
\caption{Compact test-time feedback variants for MulFeRL, GRPO, and Critique-GRPO. Variant names are defined in the text; all feedback-conditioned variants use one round of feedback-guided regeneration at test time. Values are 5-run means, with standard deviations reported for averages.}
\label{tab:feedback_disentangle_panelB}
\end{table*}

\subsection{Training Dynamics and Compute}
\label{app:extra:training_dynamics}

\subsubsection{Convergence and Validation Curves}
\label{app:extra:convergence}

\begin{figure*}[t]
  \centering
  \begin{subfigure}[t]{0.49\textwidth}
    \centering
    \includegraphics[width=\linewidth]{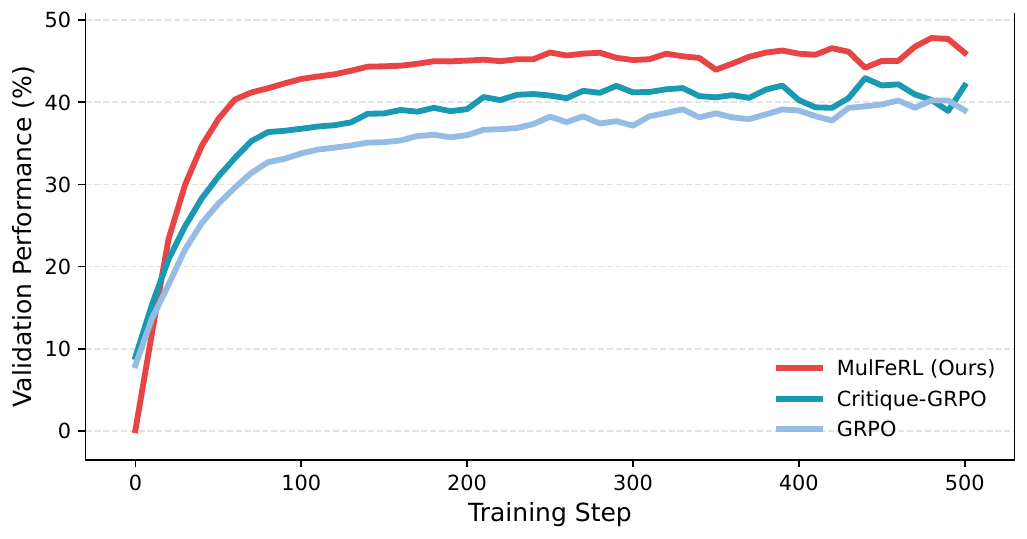}
    \caption{Qwen2.5-7B-Base.}
    \label{fig:main_a}
  \end{subfigure}\hfill
  \begin{subfigure}[t]{0.49\textwidth}
    \centering
    \includegraphics[width=\linewidth]{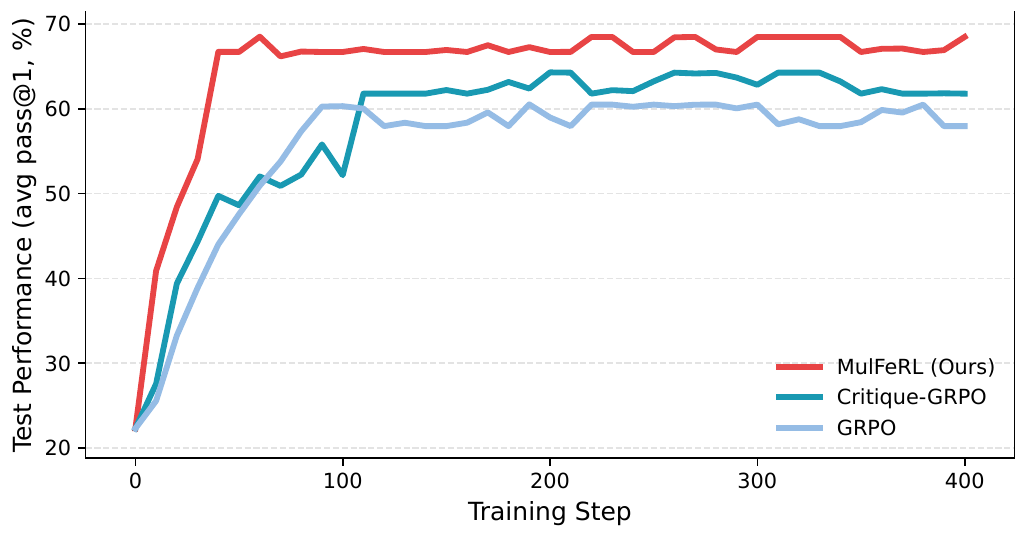}
    \caption{Qwen3-4B-Inst.}
    \label{fig:main_b}
  \end{subfigure}
\caption{Average Pass@1 over the five mathematical reasoning benchmarks across training checkpoints. MulFeRL is evaluated on both answer correctness and format compliance, while GRPO and Critique-GRPO are evaluated on answer correctness only.}
  \label{fig:con_two_subfigs}
\end{figure*}

Figure~\ref{fig:con_two_subfigs} plots the average performance on five math benchmarks as training progresses. We observe that MulFeRL improves more rapidly in the early stage and stabilizes at a strong level in substantially fewer optimization steps than GRPO and Critique-GRPO across both backbones. We attribute this faster convergence to two design choices. First, feedback-guided regeneration can often refine otherwise unsuccessful rollouts into higher-quality candidates before parameter updates, thereby reducing the extent to which the learning signal is dominated by low-quality samples. Second, cross-state FCO extracts additional supervision from verifier-certified feedback-induced transitions, providing denser and more informative learning signals per prompt.

\end{document}